\documentclass{article}
\usepackage{ijcai24}

\usepackage{times}
\usepackage{soul}
\usepackage{url}
\usepackage[hidelinks]{hyperref}
\usepackage[utf8]{inputenc}
\usepackage[small]{caption}
\usepackage{graphicx}
\usepackage{amsmath}
\usepackage{amsthm}
\usepackage{booktabs}
\usepackage{algorithm}
\usepackage[switch]{lineno}

\usepackage{subfig}

\usepackage{amsmath}
\usepackage{amssymb}
\usepackage{mathtools}
\usepackage{amsthm}

\usepackage{algorithm}
\usepackage{algpseudocode}

\usepackage{multirow}
\usepackage{adjustbox}
\usepackage{caption}
\usepackage{tabularx}

\usepackage{xcolor}

\newtheorem{theorem}{Theorem}

\title{Towards Dynamic Trend Filtering through Trend Point Detection \\ with Reinforcement Learning}

\author{
Jihyeon Seong $^1$
\and
Sekwang Oh $^1$\And
Jaesik Choi $^{1,2}$\\
\affiliations
$^1$Korea Advanced Institute of Science and Technology (KAIST), South Korea \\
$^2$INEEJI, South Korea\\
\emails
\{jihyeon.seong, oskoskosk, jaesik.choi\}@kaist.ac.kr,
}
\begin{document}

\maketitle

\begin{abstract}
    Trend filtering simplifies complex time series data by applying smoothness to filter out noise while emphasizing proximity to the original data. However, existing trend filtering methods fail to reflect abrupt changes in the trend due to `approximateness,' resulting in constant smoothness. This approximateness uniformly filters out the tail distribution of time series data, characterized by extreme values, including both abrupt changes and noise. In this paper, we propose Trend Point Detection formulated as a Markov Decision Process (MDP), a novel approach to identifying essential points that should be reflected in the trend, departing from approximations. We term these essential points as Dynamic Trend Points (DTPs) and extract trends by interpolating them. To identify DTPs, we utilize Reinforcement Learning (RL) within a discrete action space and a forecasting sum-of-squares loss function as a reward, referred to as the Dynamic Trend Filtering network (DTF-net). DTF-net integrates flexible noise filtering, preserving critical original subsequences while removing noise as required for other subsequences. We demonstrate that DTF-net excels at capturing abrupt changes compared to other trend filtering algorithms and enhances forecasting performance, as abrupt changes are predicted rather than smoothed out.
\end{abstract}

\section{Introduction}
Trend filtering emphasizes proximity to the original time series data while filtering out noise through smoothness \cite{leser1961simple}. Smoothness in trend filtering simplifies complex patterns within noisy and non-stationary time series data, making it effective for forecasting and anomaly detection \cite{Park_2020}. While smoothness achieves the property of noise filtering, an `abrupt change' denotes a point in a time series where the trend experiences a sharp transition, signaling a change in slope. Given that abrupt changes determine the direction and persistence of the slope, it is crucial to incorporate them into the trend. Traditional trend filtering employs a sum-of-squares function to reflect abrupt changes while utilizing second-order differences as a regularization term to attain smoothness \cite{hodrick1997postwar,kim2009ell_1}. However, we found that the constant nature of smoothness filters out abrupt changes, making it challenging to distinguish them from noise.

The issue of constant smoothness arises from the reliance on the property of `approximateness.' Evidence presented by \cite{ding2019modeling} suggests that the sum-of-squares function eliminates tail distribution as outliers since it approximates a Gaussian distribution with a light-tail shape. As both abrupt changes and noise reside within the tail distribution, filtering out only noise becomes challenging. This uniform filtering results in the loss of valuable abrupt changes that should be reflected in the trend \cite{ijcai2019p535}.

In this paper, we propose Trend Point Detection formulated as a Markov Decision Process (MDP), aiming to identify essential points that should be reflected in the trend, departing from approximateness \cite{sutton2018reinforcement}. These essential points are termed Dynamic Trend Points (DTPs), and trends are extracted by interpolating them. We utilize the Reinforcement Learning (RL) algorithm within a discrete action space to solve the MDP problem, referred to as a Dynamic Trend Filtering network (DTF-net) \cite{schulman2017proximal}. RL can directly detect essential points through an agent without being constrained by fixed window sizes or frequencies within the time series data domain. This dynamic approach enables the adjustment of noise filtering levels for each sub-sequence within the time series.

Building on prior research regarding reward function learning based on Gaussian Process (GP) \cite{biyik2020active}, we define the reward function as the sum-of-squares loss function from Time Series Forecasting (TSF). This choice is supported by \cite{ding2019modeling}, which suggests that the sum-of-squares function approximates a Gaussian distribution and functions similarly to a Gaussian kernel. Note that using a Gaussian kernel function as a reward leverages RL to effectively optimize the agent while learning the full distribution of time series data. Through the TSF reward, temporal dependencies around DTPs can be captured, and the level of smoothness is controlled by adjusting the forecasting window size. Additionally, to address the overfitting issue, we apply a random sampling method to both the state and the reward. 

We compare DTF-net with four categorized baselines: trend filtering (TF), change point detection (CPD), anomaly detection (AD), and time series forecasting (TSF) algorithms. First, traditional TF approaches commonly rely on approximations achieved through optimizing sum-of-squares functions or employing decomposition methods, which often neglect abrupt changes as noise. Second, CPD methods are rooted in probabilistic frameworks, prioritizing the detection of changes in distribution while often disregarding extreme values as outliers. Third, AD methods concentrate heavily on identifying abnormal points, sometimes overlooking the significance of distribution shifts in the data. Lastly, TSF models are categorized into decomposition-based and patching-based models, which also neglect abrupt changes. Contrary to all the aforementioned baselines, DTF-net focuses on point detection to reflect abrupt changes in the trend, enhancing the performance of trend filtering and forecasting. To the best of our knowledge, this is the first approach that employs MDP and RL for trend filtering, aiming to reflect both abrupt changes and smoothness simultaneously.

\begin{figure}
    \centering
    {\includegraphics[width=0.45\textwidth]{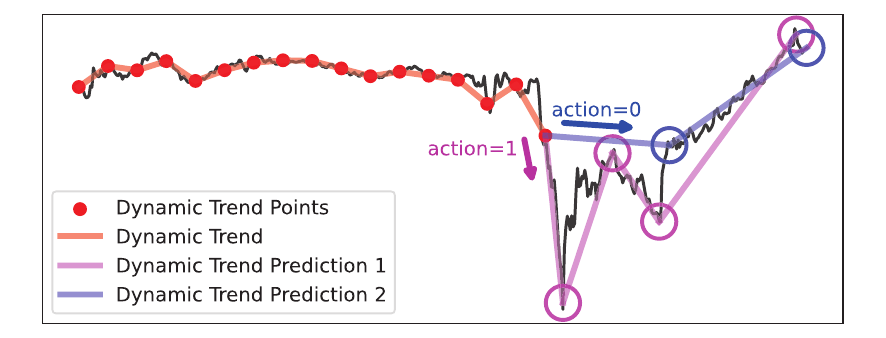}}
    \caption{\textbf{Dynamic Trend Filtering.} DTF-net extracts dynamic trends from time series data. Dynamic Trend Points (DTPs) are determined based on action predictions, and the dynamic trend is extracted through interpolation. The agent's action prediction directly influences the variation in trend extraction.}
    \label{fig:mesh1}
\end{figure}

Therefore, our contributions are as follows:

\begin{itemize}
\item We identified the issue of `approximateness,' which leads to constant smoothness in traditional trend filtering, filtering out both abrupt changes and noise.
\item We introduce Trend Point Detection formulated as an MDP, aiming to identify essential trend points that should be reflected in the trend, including abrupt changes. Additionally, we propose DTF-net, an RL algorithm that predicts DTPs through agents.
\item We employ the forecasting sum-of-squares cost function, inspired by reward function learning based on GP, which allows for the consideration of temporal dependencies when capturing DTPs. A sampling method is applied to prevent the overfitting issue.
\item We demonstrate that DTF-net excels at capturing abrupt changes compared to other trend filtering methods and enhances performance in forecasting tasks.
\end{itemize}

\section{Related Work}
\subsection{Trend Filtering}
\label{sec2.1}
Traditional trend-filtering algorithms have employed various methods to capture abrupt changes. H-P \cite{hodrick1997postwar} and $\ell_1$ \cite{kim2009ell_1} optimize the sum-of-squares function, a widely used cost function for trend filtering. However, they often face challenges in the delayed detection of abrupt changes due to the use of second-order difference operators for smoothness. To address this issue, the TV-denoising algorithm \cite{chan2001digital} was introduced, relying on first-order differences. Nevertheless, this strategy introduces delays in detecting slow-varying trends while overly focusing on abrupt changes. These methods encounter difficulties in handling heavy-tailed distributions due to the use of the sum-of-squares function \cite{ijcai2019p535}.

Contrary to sum-of-squares function methods, alternative approaches to trend filtering exist. For example, frequency-based methods like Wavelet \cite{craigmile2002wavelet} are designed for non-stationary signals but are susceptible to overfitting. The Empirical Mode Decomposition (EMD) algorithm \cite{wu2007trend} decomposes a time series into a finite set of oscillatory modes, but it generates overly smooth trends. Lastly, the Median filter \cite{siegel1982robust} is a non-linear filter that selects the middle value from the sorted central neighbors; therefore, outlier values that deviate significantly from the center of the data are excluded.

\subsection{Extreme Value Theorem}
Abrupt changes in a time series reside in the tail of the data distribution, making them rare events. However, their impact is significant, as they can alter the slope of the time series and affect the consistency of trends. Once an abrupt change occurs, its effects are often permanent until the next one occurs. Therefore, detecting abrupt changes is crucial to minimize false negative rates and capture important information.

Real-world time series data commonly exhibit a long-heavy tail distribution. Formally, the tail distribution is defined as follows:
\begin{equation}
\text{lim}_{T \rightarrow \infty} P\{ max(y_1, ..., y_T) \leq y \} = \text{lim}_{T \rightarrow \infty} F^T (y) = 0,
\label{eq:extreme}
\end{equation}
where $T$ random variables $\{y_1, ... , y_T\}$ are i.i.d. sampled from distribution $F_Y$ \cite{von1921variationsbreite,ding2019modeling}. Furthermore, extreme values within the tail distribution can be modeled using Extreme Value Theory.

\begin{theorem}[Extreme Value Theory \cite{fisher1928limiting,ding2019modeling}]
\label{theo}
If the distribution in Equation (\ref{eq:extreme}) is not degenerate to 0 under a linear transformation of $y$, the distribution of the class with the non-degenerate distribution $G(y)$ should be as follows:
\begin{equation}
G(y) = 
    \begin{cases}
    exp(- (1- \frac{1}{\gamma} y) ^\gamma),  & \gamma \neq 0, 1-\frac{1}{\gamma} y \ge 0, \\
    exp( -e^{-y}), & \gamma=0.
    \end{cases}
\end{equation}
\label{evt}
\end{theorem}

\begin{figure*}
    \centering
    \subfloat[DTF-net Environment]{\includegraphics[width=0.3\textwidth]{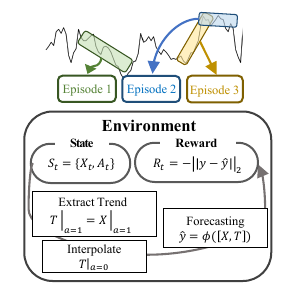}}
    \subfloat[DTF-net Reward]{\includegraphics[width=0.7\textwidth]{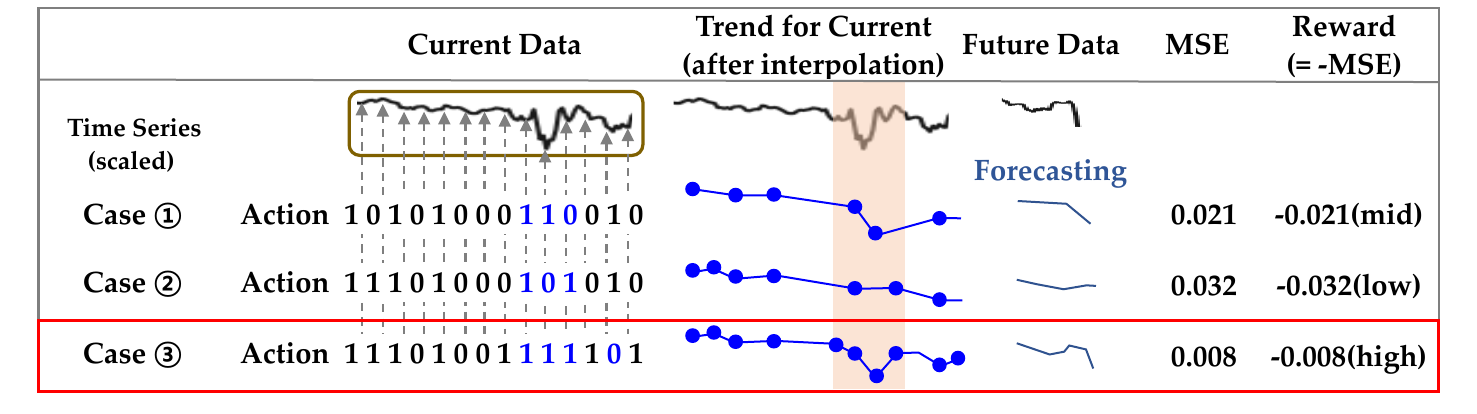}}
    \caption{\textbf{DTF-net Architecture.} DTF-net has three processes to detect DTPs: 1) The agent predicts actions within a discrete space; 2) With the predicted actions, trends are extracted by interpolating them; 3) The agent is updated through the forecasting sum-of-squares function as a reward; with time series data $X$ and trend $\mathcal{T}$ as inputs. For the reward calculation, as demonstrated in (b)-Case 3, when DTF-net successfully identifies abrupt changes, the prediction outcomes significantly improve, resulting in the highest reward.}
    \label{fig:reward}
\end{figure*}

Extreme Value Theory (EVT) has demonstrated that extreme values exhibit a limited degree of freedom \cite{lorenz1963deterministic}. This implies that the occurrence patterns of extreme values are recursive and can be memorized by a model \cite{altmann2005recurrence}. Essentially, a model with substantial capacity and temporal invariance can effectively learn abrupt changes, which are categorized as extreme values.

However, extreme values are typically either unlabeled or imbalanced, making them challenging to predict. In classification tasks, previous research \cite{class-imbalance} has highlighted the susceptibility of deep networks to the data imbalance issue. In forecasting tasks, \cite{ding2019modeling} provided evidence that minimizing the sum-of-squares loss presupposes a Gaussian distribution, which differs significantly from long-heavy-tail distributions. Motivated by these issues, we define Trend Point Detection as a problem formulation aimed at detecting essential points that should be reflected in the trend rather than smoothed out. This formulation identifies DTPs, which encompass abrupt changes, midpoints of distribution shifts, and other critical points influencing changes in the trend slope, occurring in both short and long intervals. As illustrated in Figure \ref{fig:mesh1}, Trend Point Detection is formulated as an MDP and utilizes RL to detect abrupt changes directly through the agent's action prediction. Note that we train the DNNs as a policy network of an RL agent to learn the pattern of extreme value occurrence, distinct from approximating abrupt changes as output.

\subsection{Markov Decision Process and Reinforcement Learning}
MDP is a mathematical model for decision-making when an agent interacts with an environment. It relies on the first-order Markov property, indicating that the future state depends solely on the current state. MDP comprises components denoted as $\langle S, A, P, \mathcal{R}, \gamma \rangle$. Here, $S$ denotes the set of environment states, while $A$ represents the set of actions undertaken by the agent at state $S$. The transition probability, $P = \operatorname{Pr}(S' | S, A)$, signifies the probability of transitioning from the current state $S$ to the next state $S'$. The reward, $\mathcal{R} = \mathbb{E}[\mathcal{R}(S, A, S') | S, A]$, where $\mathcal{R}(S, A, S')$ represents the immediate reward obtained when transitioning from state $S$ to $S'$ by taking action $A$. The discount factor $\gamma \in (0, 1]$ governs the trade-off between current and future rewards \cite{sutton2018reinforcement}. We can formulate any time series data with an MDP for Trend Point Detection, as detecting points always adheres to the first-order Markov property \cite{RLAD}. These points are determined solely by the current time step and remain unaffected by past observations, sharing properties similar to those of predicting stock trading points.

In RL, actions are predicted through a policy network denoted as $\pi(A|S) = \operatorname{Pr}(A|S)$ for each state, representing the probability of action $A$ at state $S$. The state-value function $v_{\pi}(S)=\mathrm{E}_{\pi}[G|S]$ estimates the expected reward value for a state $S$ under policy $\pi$, where $G=\sum_{k=0}^{\infty} \gamma^k \mathcal{R}'_{k}$ denotes the expected sum of future rewards starting from the next reward $R'$. In RL of discrete action spaces, methods like Advantage Actor-Critic (A2C) \cite{A2C} and Proximal Policy Optimization (PPO) \cite{schulman2017proximal} directly train the policy $\pi$ using the estimated state-value function $v$. In contrast, Deep Q-Network (DQN) \cite{mnih2015human} finds the optimal action-value function, denoted as $q_\pi(S,A)= \mathbb{E}_\pi[G|S,A]$. This function represents the expected cumulative reward for taking action $A$ in state $S$ under policy $\pi$ and is determined through the Bellman equation (Appendix \ref{app: RL related work}). DTF-net utilizes RL to extract flexible trends through dynamic action prediction from a deep policy network $\pi$, learning within the time series data environment formulated as an MDP of the Trend Point Detection problem.

\section{Dynamic Trend Filtering Network}
\subsection{Trend Point Detection}

\subsubsection{Environment Definition}
Time series data is defined as $\mathbf{T} = \{(\mathbf{X}_1, y_1), (\mathbf{X}_2, y_2), \ldots, \\ (\mathbf{X}_N, y_N)\}$, where $\mathbf{X}\in \mathbb{R}^D$ represents the input, $y\in \mathbb{R}^d$ represents the output, and the dataset comprises a total of $N \in \mathbb{Z}^+$ samples. Here, $D$ and $d$ denote the input and output dimensions, respectively, both of which are positive integers.

The Trend Point Detection problem formulation takes input $\mathbf{X}$ representing the environment and outputs DTPs, encompassing abrupt changes, midpoints of distribution shifts, and other critical points influencing trend slope changes occurring at both short and long intervals. The output consists of specific univariate time series $y^{(i)}$ labeled with binary values, where $i \in d$ of target.

\begin{itemize}
\item \textbf{State $S = [\mathbf{X}_{t}, A_{t}]$}: the positional encoded vector set of time series data $\mathbf{X}$ and action $A$ with horizon $t$.
\item \textbf{Action $A$}: a discrete set with $(a=1)$ for detecting DTP and $(a=0)$ for smoothing.
\item \textbf{Reward $\mathcal{R}(S, A, S')$}: the change in forecasting sum-of-squares function value when action $A$ is taken at state $S$ and results in the transition to the next state $S'$.
\item \textbf{Policy $\pi(A|S)$}: the probability distribution of $A$ at $S$.
\end{itemize}

The RL algorithm, named DTF-net, employs a policy network $\pi$ within the defined MDP. It receives the state $S$ as input and outputs the binary labeled target $y^{(i)}$, also denoted as $A$, learned through the maximization of cumulative rewards $\mathcal{R}$. DTF-net is designed to extract dynamic trends by interpolating detected essential trend points, referred to as DTPs and represented by the set $\{y^{(i)} = 1\}$ or $\{A|_{a=1}\}$.

\subsubsection{Episode and State for DTF-net}
Previous studies in RL for time series \cite{liu2022finrl_meta} have generally adopted a sequential approach. In contrast, DTF-net introduces dynamic segmentation with variable lengths comprising one episode through random sampling. The discrete uniform distribution is specifically chosen to ensure that all sub-sequences are considered equally:
\begin{equation}
    \begin{split}
        &s \sim \text{unif}\{0, N\},\\
        &l \sim \text{unif}\{h+p, H\},
    \end{split}
\end{equation}
where $s$ represents the starting points of the sub-sequence, $l$ denotes the sub-sequence length, $h$ denotes the forecasting look-back horizon, $p$ denotes the forecasting prediction horizon, and $H$ represents the maximum length comprising one episode. With sampling, the length and starting point of the sub-sequence are defined, resulting in a non-sequential and random progression of the episode. This sampling approach mitigates the overfitting issue by allowing the model to use only a portion of the sequence.

Within a single episode, DTF-net cumulatively constructs the state $S$. To maintain a constant state length within an episode, we employ positional encoding as follows:
$$PE_{(pos,2i)} = sin(pos/10000^{2i/d_{model}}),$$
$$PE_{(pos,2i+1)} = cos(pos/10000^{2i/d_{model}}).$$

The cumulative state progression is achieved by gradually expanding the state representation $S_t$ as the step unfolds as follows,
\begin{equation}
    S_t = PE(\{ \mathbf{X}_{s:s+t}, A_{0:t} \}), \text{ where } t<l.
\end{equation}
Through cumulative state construction, the agent can learn sequential information in the time series (Appendix \ref{app: state}).

\subsection{Reward Function of DTF-net}
\subsubsection{GP and Reward Function Learning}
\begin{algorithm}
\caption{Reward Procedure of DTF-net}\label{reward}
\begin{algorithmic}
\Procedure{Reward}{$S_{t}$}%\Comment{$\mathbf{X}, A \in E_{t-(h+p):t}$ at time step $t$}
    \State $\mathbf{X}' = \mathbf{X}_{t-(h+p):t}, A' = A_{t-(h+p):t}$ 
    \State $\mathcal{T} \gets 0$%\Comment{trend initialization}
   \State $\mathcal{T}|_{a=1} \gets \mathbf{X}'_{0:h}|_{a=1}$%\Comment{value assign for a=1}
   \While{$n\le h$}%\Comment{for linear interpolation}
      \State // $n$ for time-axis and $\mathbf{x} \in \mathbf{X}'$
      \State $ \mathcal{T}_n \gets \mathbf{x}_n = \mathbf{x}_{n-1} + \frac{\mathbf{x}_{n+1} - \mathbf{x}_{n-1}}{2}$
      \State $n \gets n + 1$
   \EndWhile
   \State $\hat{y} \gets \phi([\mathbf{X}'_{0:h}, \mathcal{T}])$ %\Comment{Regression for TSF}
   \State $r \gets \frac{1}{p} \sum_{i=1}^p (y_i - \hat{y}_i)^2$ %\Comment{MSE loss}
   \State \textbf{return} $-r$%\Comment{minus of forecasting reward}
\EndProcedure
\end{algorithmic}
\end{algorithm}
Traditional trend filtering methods utilize the sum-of-squares function, also known as the Mean Squared Error (MSE), to approximate abrupt changes when extracting trends. However, \cite{ding2019modeling} provided evidence that minimizing the sum-of-squares function assumes that the model output distribution determined using the MSE cost function denoted as $\hat{P}(Y)$, follows a Gaussian distribution with variance $\tau$, grounded in Bregman's theory \cite{banerjee2005clustering}.
\begin{equation}
\begin{split}
\hat{P}(Y)
    &=\text{min} \sum_{t=1}^T ||y_t - o_t||^2, \\
    &= \text{max}_\theta \Pi_{t=1}^T P(y_t | x_t, \theta),\\
    &=\frac{1}{N} \sum_{t=1}^T \mathcal{N}(y_t, \hat{\tau}^2).
\end{split}
\label{mse}
\end{equation}
where $o \in Y$ represents the output from a model parameterized by $\theta$.
This also suggests that model $\theta$ operates in a manner similar to a Kernel Density Estimator (KDE) employing a Gaussian kernel \cite{rosenblatt1956remarks}.

Contrary to approximations, DTF-net utilizes a policy network $\pi$ to predict DTPs, including abrupt changes. However, defining a reward function in general time series data is challenging but is the most crucial task in RL training for optimizing the policy network. To tackle this challenge, DTF-net draws inspiration from previous works, which employ the Gaussian Process (GP) for reward function learning \cite{nips_gaussian_reward,biyik2020active}. 

\begin{table*}
\centering
\begin{minipage}[b]{0.7\linewidth}
\resizebox{0.95\textwidth}{!}{
    \begin{tabular}[b]{c|c|cc|cc}
        \hline
         \multicolumn{2}{c|}{\multirow{3}{*}{Trend Filtering}} & \multicolumn{4}{c}{Linear Signal+Noise (0.2)}  \\
         \cline{3-6}
         \multicolumn{2}{c|}{} & \multicolumn{2}{c|}{1) full-sequence} & \multicolumn{2}{c}{2) abrupt-sequence}\\
        \cline{3-6}
         \multicolumn{2}{c|}{} & MSE & MAE & MSE & MAE \\
        \hline
             \multirow{2}{*}{CPD} &ADAGA \cite{adaga} & 4.3434 & 1.4428 & 7.0120 & 1.8668\\
             &RED-SDS \cite{ansari2021deep} & 1.0036 & 0.6782 &  1.6660 & 0.9365\\
             \hline
             \multirow{3}{*}{AD} &TimesNet \cite{wu2023timesnet} & 3.0841 & 1.4204 & 3.3304 & 1.4364 \\
             &AnomalyTransformer \cite{xu2022anomaly} & 7.7506 & 2.1817 & 10.1336 & 2.7242\\
             &DCdetector \cite{yang2023dcdetector} & \underline{\textit{0.0094}} & \underline{\textit{0.0255}} & 3.6300 & 1.3721\\
             \hline
             \multirow{7}{*}{TF} &EMD \cite{wu2007trend} & 5.3096 & 1.7401 & 6.4410 & 1.8431 \\
             &Median \cite{siegel1982robust} & 4.4766 & 1.5525 & 5.6859 & 1.8204\\
             &H-P \cite{hodrick1997postwar} & 0.2253 & 0.3311 & 0.3238 & 0.3934\\
             &Wavelet \cite{craigmile2002wavelet} & $1e-30$ & $6e-16$ & $2e-30$ & $8e-16$\\
             &$\ell_1$ ($\lambda$=0.1) \cite{kim2009ell_1}& \underline{0.0461} & \underline{0.1703} & \underline{0.0500} & \underline{0.1807}\\
             &$\ell_1$ ($\lambda = 5e-4$) & 0.0004 & 0.0175 & 0.0004 & 0.0174\\
             \cline{2-6}
              &DTF-net (ours) & \bf 0.0289 & \bf 0.0826 & \bf 0.0286 & \bf 0.0855\\
        \hline
      \end{tabular}}
\end{minipage}
      \caption{\textbf{Comparison with advanced CPD, AD, and TF methods in synthetic data.} We conduct trend filtering analysis on synthetic data, evaluating it against the ground truth of a linear signal with added noise. We consider two cases: one with the full sequence and the other with a 30-window interval sub-sequence containing abrupt changes. The evaluation metrics are Mean Squared Error (MSE) and Mean Absolute Error (MAE), where lower values indicate better performance. The best performance is \textbf{bolded}, and the second-best performance is \underline{underlined}. For the special case of DCdetector, the performance is denoted in \textit{\underline{italic}}.}
      \label{tab: syn}
\end{table*}

Formally, GP \cite{gaussian_process} assumes noisy targets $y_i=f(x_i) + \epsilon_i$ that are jointly Gaussian with a covariance function $k$:
\begin{equation}
    P(y|x) \sim \mathcal{N}(0, \mathbf{K}), \text{ where } \mathbf{K}_{pq} = k(x_p, x_q).
\end{equation}
With a Gaussian covariance function, 
$$k(x_p, x_q | \theta) = v^2 exp(-(x_p-x_q)^\top \Lambda^{-1} (x_p - x_q)/2) + \delta_{pq} \sigma^2_n, $$
where diagonal matrix $\Lambda$, $v$, and $\sigma$ are hyperparameters in $\theta$, the predictive distribution for input $x^*$ follows Gaussian: 
\begin{equation} 
    \begin{split}
        P(o^*_t | x^*, x, y, \theta) \sim &\mathcal{N}(k(x^*, x)\mathbf{K}^{-1}y, \\
         & k(x^*, x^*)-k(x^*,x) \mathbf{K}^{-1}k(x, x^*)).
    \end{split}
    \label{gp}
\end{equation}

The GP model inherently learns a full distribution of time series data, enabling RL to effectively optimize the policy network (Appendix \ref{app: RL related work}). Leveraging these insights, DTF-net's reward function is defined as the sum-of-squares function from Time Series Forecasting (TSF). This choice leads to more efficiency in calculating rewards compared to GP while achieving reward function learning within the Gaussian distribution. To incorporate captured abrupt changes into the forecasting model, DTPs are included as an additional input. As shown in Figure \ref{fig:reward}, when the forecasting model predicts upward or downward trends instead of smoothing them out, the agent receives a higher reward. Thus, DTF-net learns temporal dependencies when capturing DTPs.

\subsubsection{Forecasting Reward Function of DTF-net}
As shown in Algorithm \ref{reward}, the reward process involves time series data $\mathbf{X}_{t-(h+p):t}$ and action $A_{t-(h+p):t}$ in state $S_{t}$ at time step $t$, both having a sequence length denoted as $(h+p)$, where $h$ denotes the past horizon and $p$ denotes the forecasting horizon (under the condition $t-(h+p) > 0$). The trend $\mathcal{T}$ initiates with $\mathbf{X}$ values assigned only under the condition of action $A|_{a=1}$, and linear interpolation is applied for the remaining values. Subsequently, forecasting is conducted with a prediction length $p$ defined by a hyperparameter. The reward is computed as the negative sum-of-squares loss between the predicted $\hat{y}$ and the ground truth $y$.

DTF-net uses a penalty reward as a negative value from the Mean Squared Error (MSE) function, and there is a possibility of an overfitting issue. Therefore, DTF-net utilizes random sampling from a discrete uniform distribution, providing better control over model updates. 
\begin{equation}
    \begin{split}
        &k \sim \text{unif}\{s, s+l\},\\
        &R = 
        \begin{cases}
            \text{REWARD}(E_{t}) & \text{if } t=k,\\
            0 & \text{if } t \neq k.
        \end{cases}
    \end{split}
\end{equation}
We empirically demonstrate that irregularly applying penalties through sampling can prevent overfitting rather than penalizing at every step in Section \ref{sec4.2.3} (Appendix \ref{app: reward}).

\begin{figure}[h]
    \centering
    \includegraphics[width=0.3\textwidth]{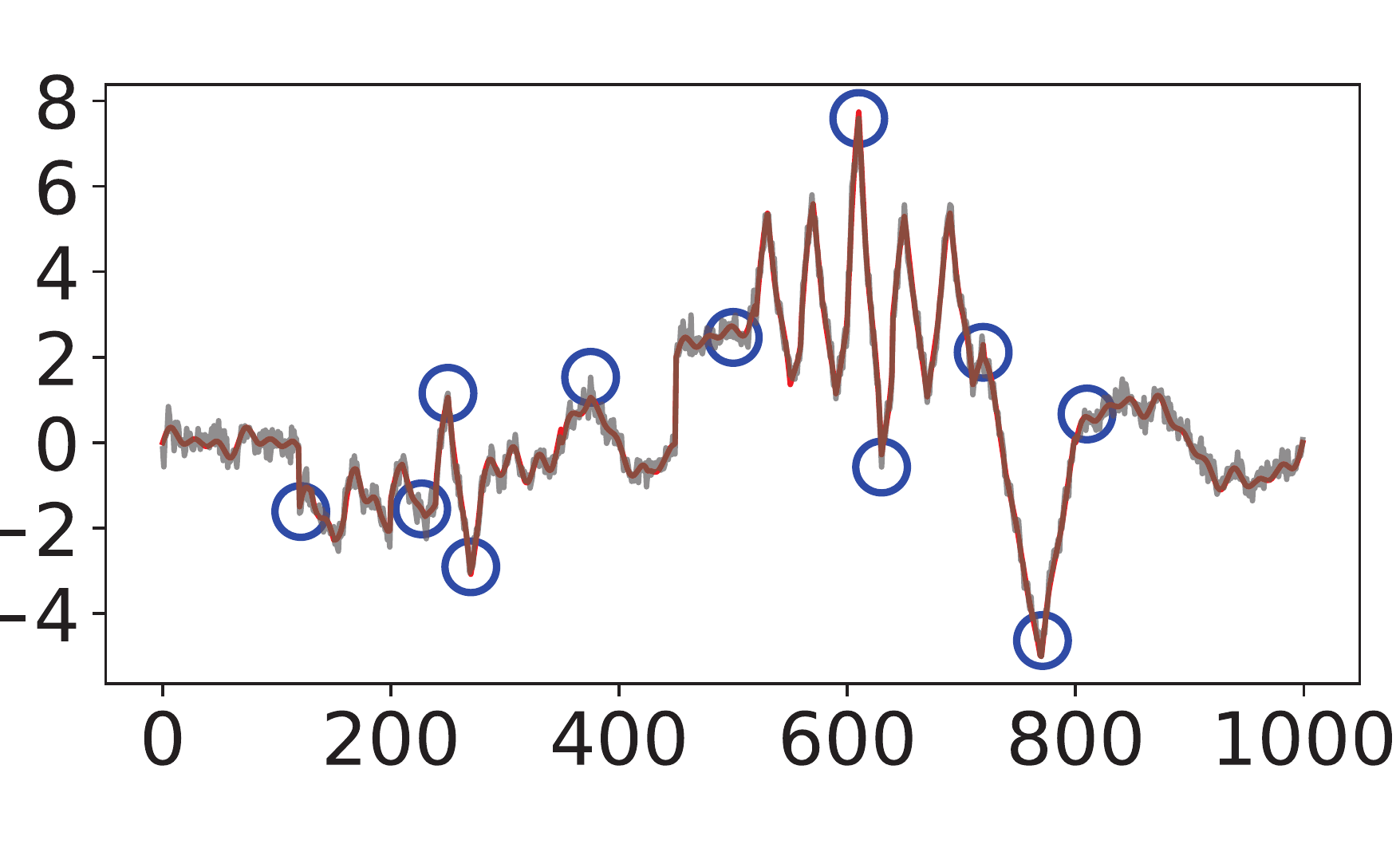}
    \caption{\textbf{Synthetic Data.}}
    \label{fig:synthetic}
\end{figure}

In summary, DTF-net is designed to extract dynamic trends $\mathcal{T}$ by interpolating detected DTPs. A simple ML time series predictor $\phi$ is integrated into DTF-net to calculate the reward, with the input of $[\mathbf{X}, \mathcal{T}]$. As shown in Figure \ref{fig:reward}, including the detected abrupt changes as an additional input to the forecasting model ensures that the forecasting output reflects both upward and downward trends, maximizing the reward. 

\section{Experiment}
\subsection{Trend Filtering Analysis}
\subsubsection{Experimental Settings}
\label{sec:exp setting}
Analyzing trend filtering methods quantitatively poses two challenges: 1) defining a ground truth for the trend is challenging, and 2) labeling abrupt changes is challenging. To address these, as shown in Figure \ref{fig:synthetic}, we generate a synthetic trend signal with $1,000$ time points. This synthetic dataset contains $11$ abrupt changes, including 1) a sudden drop to negative values around -2 and -5 at time points 100 and 800, respectively; 2) a mean shift from 0 to 4 with high variance occurring between time points 500 and 700; and 3) a sine wave starting from time point 800 to 1000, completing one cycle. We add Gaussian noise with a standard deviation of $0.2$ to simulate real-world conditions.

\begin{figure*}[h]
\centering
    \subfloat[$\ell_1 (\lambda=0.1)$]{\includegraphics[clip, trim= 8cm 2cm 8cm 2cm, width=0.33\textwidth]{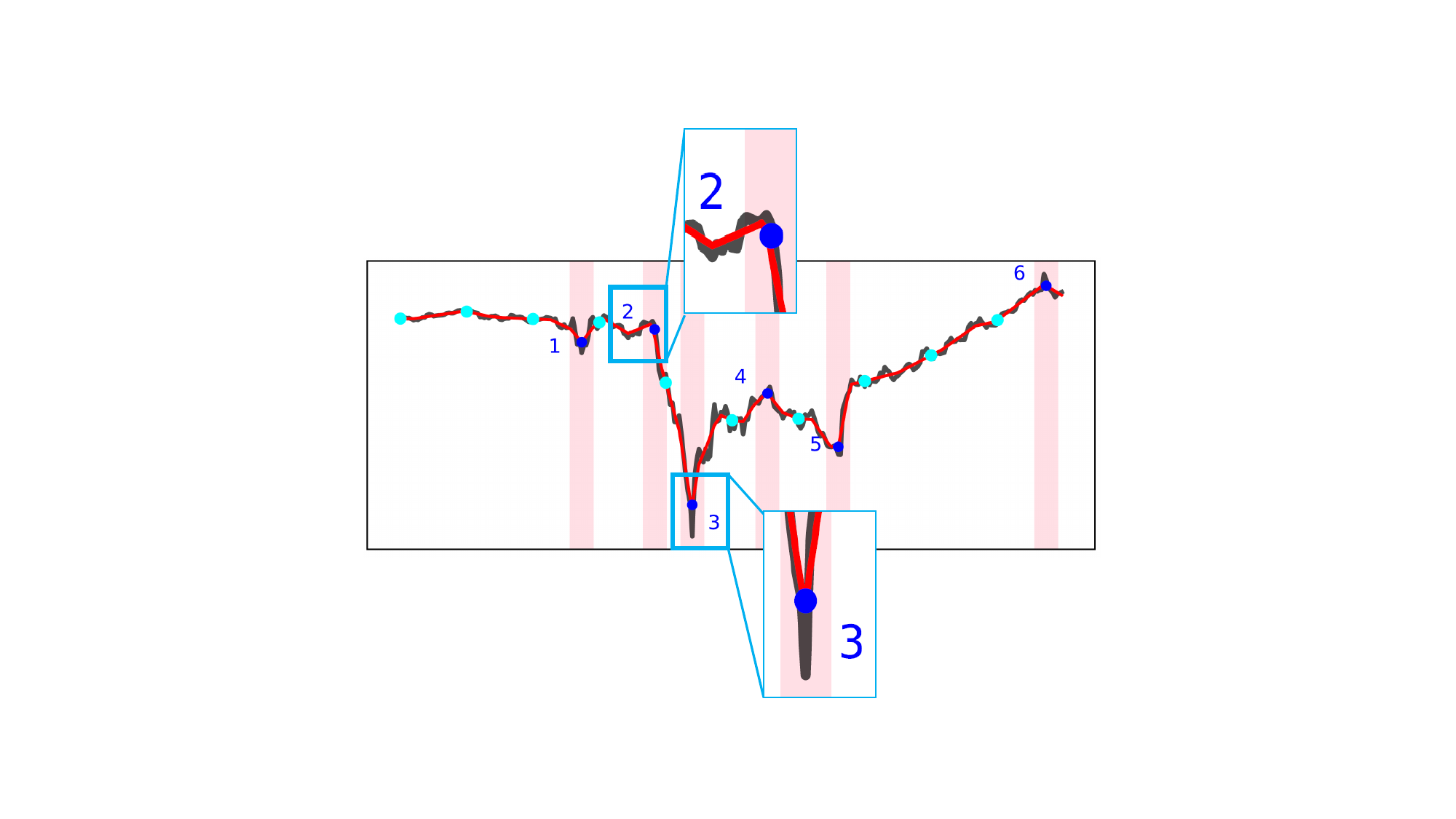}}
    \subfloat[$\ell_1 (\lambda=5e-4)$]{\includegraphics[clip, trim= 8cm 2cm 8cm 2cm, width=0.33\textwidth]{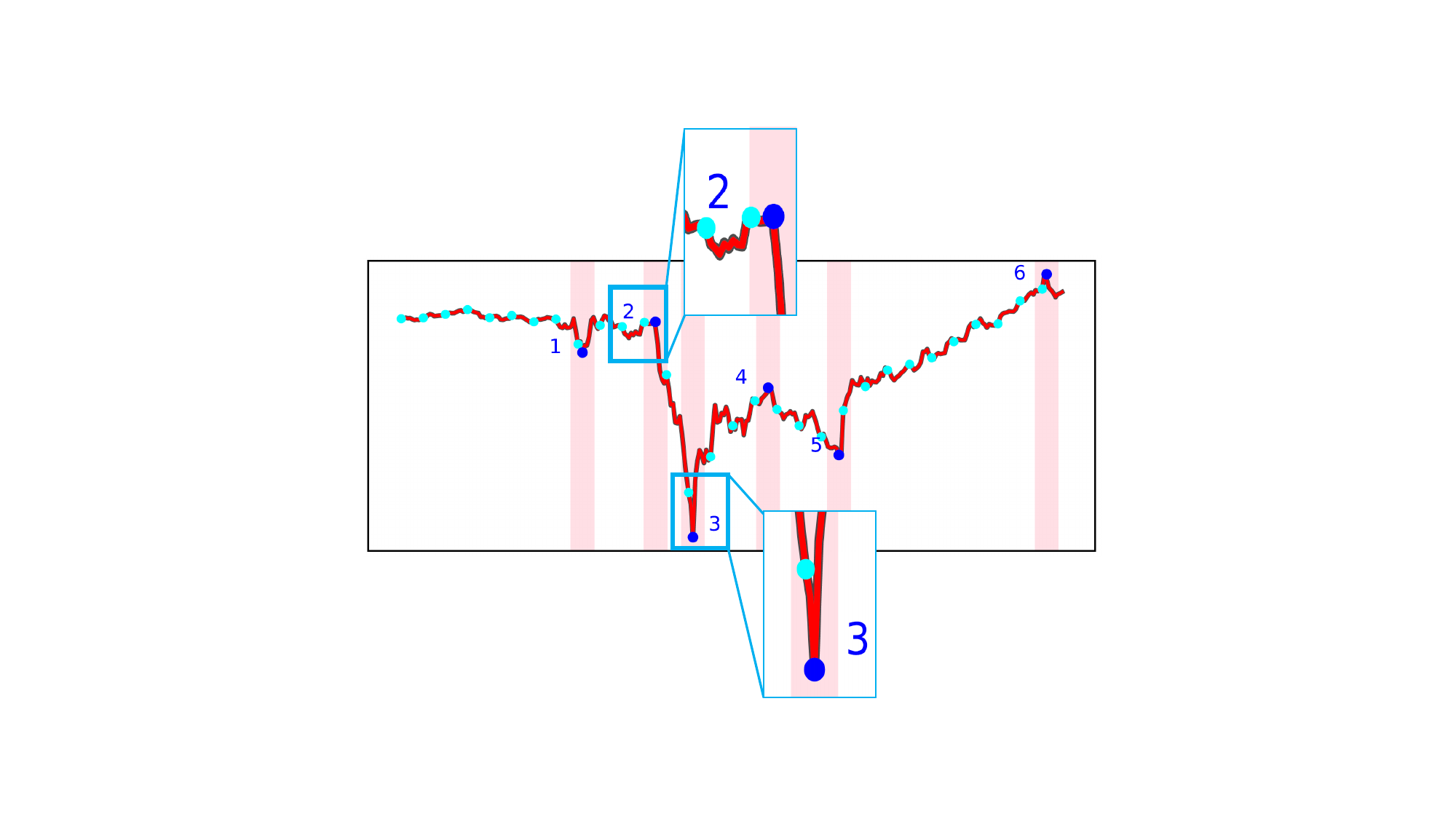}}
    \subfloat[DTF-net]{\includegraphics[clip, trim= 8cm 2cm 8cm 2cm, width=0.33\textwidth]{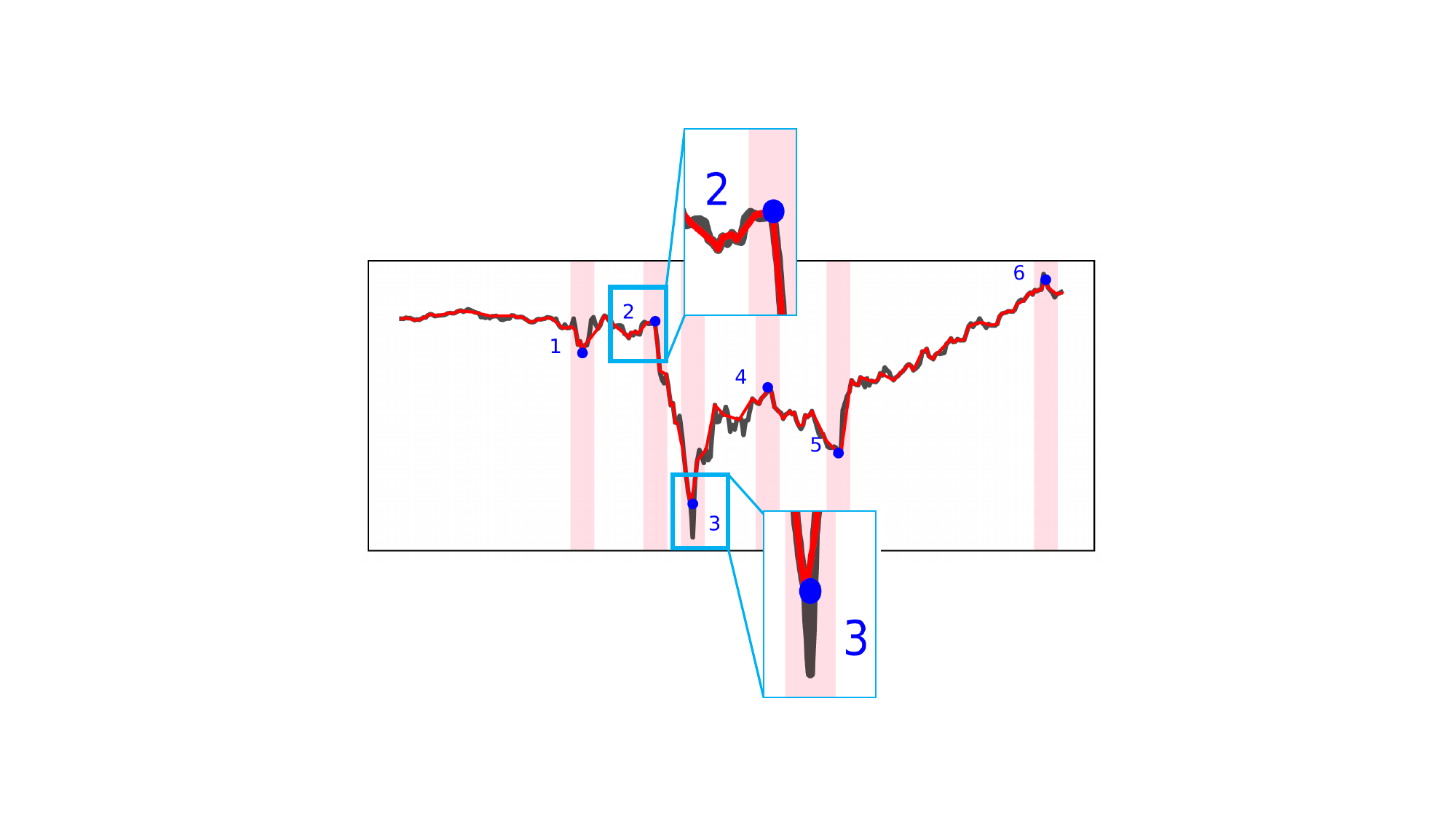}}
    \caption{\textbf{Qualitative comparison with $\ell_1$ and DTF-net.} The figure illustrates the trends obtained from the $\ell_1$ and DTF-net using the Nasdaq intraday dataset. The red line denotes the output of each trend filtering method, with red vertical boxes indicating arbitrarily set abrupt changes. The blue dots denote the captured abrupt changes, while the sky-blue dots highlight the constant smoothness from $\ell_1$. Notably, DTF-net has the capability to apply varying levels of smoothness to individual sub-sequences.}
    \label{fig:nasdaq}
\end{figure*}

\begin{table*}[t]
    \centering
    \begin{adjustbox}{width=1\textwidth}
    \begin{tabular}{c|c|cc|cc|cc|cc|cc|cc|cc|cc}  
    \hline \multicolumn{2}{c|}{Methods} & \multicolumn{2}{c|}{ DTF-Linear (ours) } & \multicolumn{2}{c|}{ $\ell_1 (\lambda =0.1)$-Linear } & \multicolumn{2}{c|}{ PatchTST/42 } & \multicolumn{2}{c|}{ NLinear } & \multicolumn{2}{c|}{ DLinear } & \multicolumn{2}{c|}{ FEDformer-f } & \multicolumn{2}{c|}{ FEDformer-w } & \multicolumn{2}{c}{ Autoformer } \\
    \hline \multicolumn{2}{c|}{Metric} & MSE & MAE & MSE & MAE & MSE & MAE & MSE & MAE & MSE & MAE & MSE & MAE & MSE & MAE & MSE & MAE \\
    \hline 
    % \cline{3-8}
    \multirow{6}{*}{Exchange} & 24 & \bf 0.0250 & \bf 0.1198 & 0.0266 & 0.1248 & 0.0387 & 0.1513 & \underline{$0.0275$} & \underline{$0.1264$} & $0.0290$ & $0.1284$ & $0.0381$ & $0.1545$ & $0.0387$ & $0.1564$ & $0.0687$ & $0.2041$ \\
    & 48 & \bf 0.0487 & \bf 0.1658 & $0.0505$ & $0.1708$ & 0.0624 & 0.1873 & \underline{$0.0505$} & \underline{$0.1705$} & $0.0585$ & $0.1907$ & $0.0548$ & $0.1818$ & $0.1068$ & $0.2528$ & $0.1095$ & $0.2485$  \\
    & 96 & \bf 0.0980 & \bf 0.2349 & 0.1007 & 0.2440 & 0.1833 & 0.3436 & \underline{$0.0990$} & \underline{$0.2361$} & $0.1063$ & $0.2530$ & $0.1440$ & $0.2980$ & $0.1386$ & $0.2894$ & $0.1834$ & $0.3306$  \\
    & 192 & 0.1983 & 0.3583 & $0.2045$ & $0.3518$ & 0.2550 & 0.3987 & $0.2030$ & \underline{$0.3400$} & \underline{$0.1959$} & $0.3554$ & $0.2790$ & $0.4163$ & $0.2841$ & $0.4217$ & $0.3465$ & $0.4510$ \\
    & 336 & \bf 0.3160 & \bf 0.4561 & $0.3337$ & $0.4666$ & 0.5161 & 0.5442 & $0.4174$ & $0.4857$ & \underline{$0.3276$} & \underline{$0.4627$} & $0.4466$ & $0.5130$ & $0.5685$ & $0.5890$ & $0.4488$ & $0.5291$ \\
    & 720 & \bf 0.7933 & \bf 0.6874 & $0.9515$ & $0.7636$ & 1.1143 & 0.8063 & $1.0420$ & $0.7807$ & \underline{$0.9071$} & \underline{$0.7415$} & $1.2122$ & $0.8492$ & $1.2912$ & $0.8876$ & $1.2463$ & $0.8694$  \\
    \hline
    \multirow{6}{*}{ETTh1} & 24 & $0.0253$ & $0.1205$ & \underline{\textit{0.0234}} & \underline{\textit{0.1140}} & \underline{0.0266} & \underline{0.1238} & $0.0266$ & $0.1240$ & $0.0273$ & $0.1262$ & $0.0358$ & $0.1450$ & $0.0381$ & $0.1524$ & $0.0694$ & $0.2042$ \\
    & 48 & $0.0375$ & $0.1479$ & \underline{\textit{0.0366}} & \underline{\textit{0.1442}} & 0.0393 & 0.1506 & \underline{$0.0388$} & \underline{$0.1503$} & $0.0404$ & $0.1523$ & $0.0547$ & $0.1778$ & $0.0602$ & $0.1921$ & $0.0797$ & $0.2205$  \\
    & 96 & \bf 0.0519 & \bf 0.1740 & 0.0521 & 0.1744 & 0.0550 & 0.1790 & \underline{$0.0519$} & \underline{$0.1745$} & $0.0551$ & $0.1815$ & $0.0786$ & $0.2126$ & $0.0919$ & $0.2348$ & $0.0857$ & $0.2292$  \\
    & 192 & \bf 0.0676 & \bf 0.2013 & $0.0693$ & $0.2034$ & 0.0705 & 0.2050 & \underline{$0.0694$} & \underline{$0.2046$} & $0.0730$ & $0.2076$ & $0.0933$ & $0.2344$ & $0.1000$ & $0.2464$ & $0.0993$ & $0.2428$ \\
    & 336 & 0.0803 & 0.2247 & \underline{\textit{0.0796}} & \underline{\textit{0.2238}} & \underline{0.0814} & \underline{0.2260} & $0.0826$ & $0.2280$ & $0.0948$ & $0.2414$ & $0.1117$ & $0.2597$ & $0.1418$ & $0.2958$ & $0.1287$ & $0.2792$ \\
    & 720 & \bf 0.0776 & \bf 0.2224 & 0.0789 & 0.2244 & 0.0869 & 0.2329 & \underline{$0.0814$} & \underline{$0.2273$} & $0.1800$ & $0.3494$ & $0.1310$ & $0.2858$ & $0.1224$ & $0.2766$ & $0.1378$ & $0.2939$  \\
    \hline
    \multirow{3}{*}{Illness} & 24 & \bf 0.5881 & \bf 0.5358 & 0.6119 & \underline{\textit{0.5299}} & \underline{0.6228} & \underline{0.5305} & $0.6325$ & $0.5639$ & $0.7831$ & $0.7462$ & $0.6969$ & $0.6256$ & $0.7100$ & $0.6352$ & $0.7432$ & $0.6704$ \\
    & 48 & \bf 0.6858 & \bf 0.6359 & $0.6925$ & $0.6322$ & 0.7109 & 0.6642 & \underline{$0.6892$} & \underline{$0.6453$} & $0.8217$ & $0.7750$ & $0.7099$ & $0.6935$ & $0.6961$ & $0.6972$ & $0.7855$ & $0.7370$  \\
    & 60 & 0.6640 & 0.6423 & 0.6666 & \underline{\textit{0.6324}} & \underline{0.6465} & 0.6381 & $0.6730$ & \underline{$0.6347$} & $0.9195$ & $0.8361$ & $0.8309$ & $0.7653$ & $0.8192$ & $0.7641$ & $0.8945$ & $0.8055$  \\
    \hline
    \end{tabular}
    \end{adjustbox}
    \caption{\textbf{Evaluating DTF-net in TSF task.} We conduct TSF experiments using three non-stationary datasets: Exchange Rate, ETTh1, and Illness. We evaluate performance using MSE and MAE, where lower values indicate better performance. In the following results, the best-performing models using DTF-net are highlighted in \textbf{bold}, and models using $\ell_1$ trend filtering are highlighted in \underline{\textit{italic}}. Additionally, for comparison, the best-performing models using only original data are \underline{underlined}.} 
    \label{tab:TSF_main}
\end{table*}

To evaluate the trend filtering results, we set up the experiment as follows. We employ Mean Squared Error (MSE) and Mean Absolute Error (MAE) metrics to measure the proximity to the original data, which is a key aspect of trend filtering. For the ground truth, we use a linear signal with added noise to assess DTF-net's robustness to noisy data. In cases with added noise, we assume that filtering out at least 10\% of noise is necessary to confirm smoothness. We divide the proximity evaluation into two categories: full-sequence and sub-sequence. In the sub-sequence evaluation, a 30-window interval is set around labeled abrupt changes to assess temporal dependencies are well captured. 

\subsubsection{Performance Analysis}
Table \ref{tab: syn} emphasizes DTF-net's superior performance compared to CPD and AD algorithms. CPD algorithms are designed to identify shifts in data distribution and tend to capture the midpoint of these changes, treating extreme values as outliers. On the other hand, AD algorithms focus exclusively on pinpointing anomalous data points, often overlooking the midpoint of changes. For instance, DCdetector is particularly adept at identifying abnormal values, demonstrating superior performance across entire data sequences. However, its effectiveness diminishes when dealing with abrupt sub-sequences. This shortfall stems from its focus solely on detecting abnormal points and short intervals surrounding abrupt changes. Consequently, while it maintains commendable performance on full sequences, it falls short in accurately filtering trends within abrupt sub-sequences.In essence, the differing objectives of CPD and AD algorithms make them less suitable for trend-filtering (Figure \ref{fig:cpd} in Appendix \ref{sec2}).

In comparison with other trend filtering methods, DTF-net outperforms all methods except those prone to overfitting. Decomposition-based methods like EMD and Median generate excessively smoothed trends. The frequency-based method, Wavelet, tends to overfit to noise. The $\ell_1$ method shows sensitivity to hyperparameter $\lambda$, as evident by $\lambda=5e-4$ causing overfitting to noise. Therefore, DTF-net excels in capturing abrupt changes while reflecting temporal dependencies within noisy and complex time series data.

\begin{figure*}[h]
    \subfloat[ETTh1 (24 pred)]{\includegraphics[clip, trim= 0 0 7.3cm 0, width=0.29\textwidth]{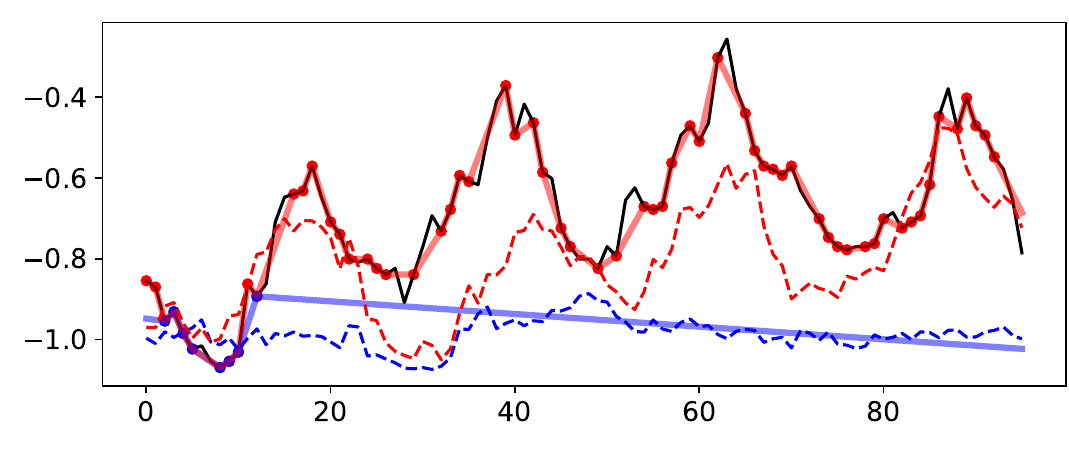}}
     \subfloat[EXC (24 pred)]{\includegraphics[clip, trim=  0cm 0 7.3cm 0, width=0.29\textwidth]{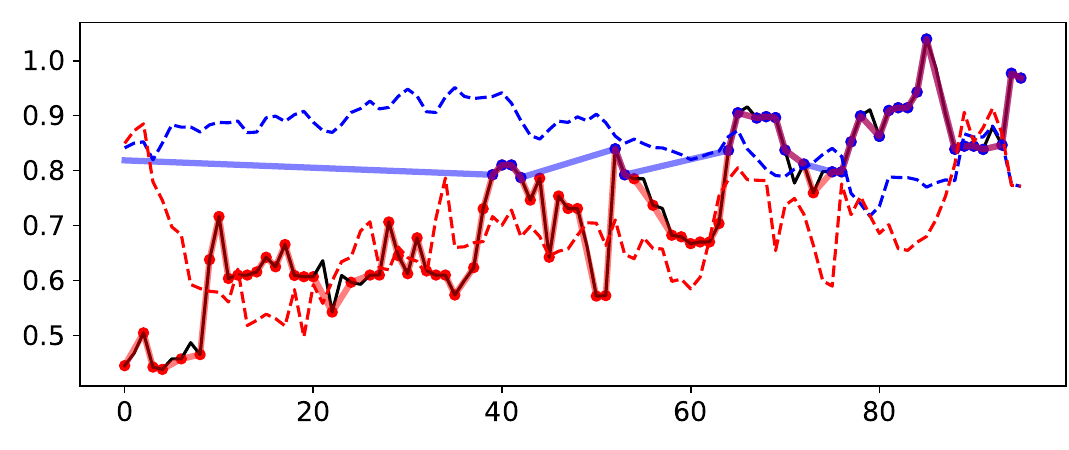}}
    \captionsetup[subfigure]{oneside,margin={0cm,2.1cm}}
    \subfloat[ETTh1 (336 pred)]{\includegraphics[clip, trim= 0cm 0 0cm 0, width=0.4\textwidth]{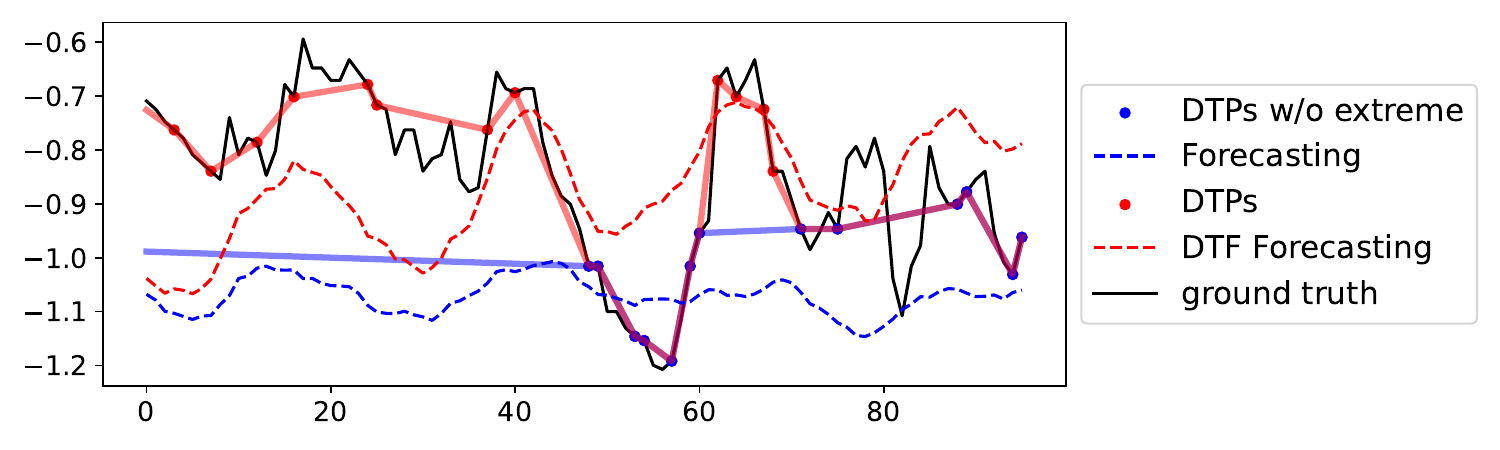}}\\

    \caption{\textbf{Qualitative analysis of the impact of abrupt changes on TSF.} We conduct forecasting experiments to evaluate the influence of trends incorporating extreme values with long-heavy tails on two datasets, ETTh1 and Exchange rate (EXC). The figure illustrates that including abrupt changes (red) in forecasting plays a crucial role without undergoing smoothing (blue). It is evident that the results appear smoother when extreme values are excluded (depicted by the blue line) in both short-term (24 pred) and long-term (336 pred) forecasting.}
    \label{fig:extreme value study}
\end{figure*}

\subsubsection{Nasdaq Dataset}
To demonstrate the proficiency of DTF-net on complex real-world datasets, we perform additional analysis on the Nasdaq intraday dataset from July 30th to August 1st, 2019, characterized by rapid changes. Here, we arbitrarily set 6 abrupt changes and qualitatively analyze the results. As shown in Figure \ref{fig:nasdaq}, it is evident that the $\ell_1$ trend filtering algorithm extracts trends that either underfit or overfit depending on the parameter $\lambda$ due to constant smoothness. For point 3, in detail, $\ell_1$ with ($\lambda=0.1$) filtered out noise, while ($\lambda=5e-4$) captured it as abrupt changes. In contrast, DTF-net accurately captures five abrupt changes and concurrently performs noise filtering for point 3. This accomplishment is attributed to the dynamic nature of trend extraction from DTF-net.

\subsection{Trend Filtering in Time Series Forecasting}
\subsubsection{Experimental Settings}
We analyze how DTF-net effectively captures abrupt changes and extend our evaluation to include a real-world dataset for a common time series task. Time Series Forecasting (TSF) models are expected to predict potential incidents associated with extreme values, providing valuable insights for critical decision-making \cite{van2008cautious}. To assess the practicality of DTF-net in real-world scenarios, we apply it to TSF, incorporating the extracted trend as an additional input feature. Formally, the forecasting model receives input as $\mathbf{X}'=[\mathbf{X}, \mathbf{P}] \in \mathbb{R}^{D+1}$, where $\mathbf{P}$ represents the trend from DTF-net. Under the same conditions, we compare this model to those using $\ell_1$ as additional inputs and only the original sequence $\mathbf{X}$ as inputs. We employ DTF-net with the TSF models NLinear and DLinear \cite{zeng2022transformers}, which are considered state-of-the-art yet simplest in the field of TSF. The experiment focuses on the univariate forecasting case to assess trend filtering effectiveness (Appendix \ref{app: TSF}). Note that DTF-net is not directly linked with TSF models; instead, the extracted trend from DTF-net is provided as additional input.

\subsubsection{Performance Analysis}
We choose three non-stationary datasets from the TSF benchmark dataset: Exchange Rate, ETTh1, and Illness (Appendix \ref{app: TSF dataset}). Table \ref{tab:TSF_main} indicates that DTF-net outperforms in most cases. Among the three datasets, the exchange rate dataset is the most intricate, exhibiting the least seasonality and the highest level of noise. Given the absence of periodicity in financial data, $\ell_1$ trend filtering encounters difficulties in extracting clear trends. However, DTF-net demonstrates robustness when dealing with non-stationary time series data.

However, models employing $\ell_1$ trend filtering have advantages when dealing with more stationary data that exhibits a recursive pattern. The piece-wise linearity assumption of $\ell_1$ is particularly pronounced in short-term predictions within ETTh1, as it is the least noisy and most stationary dataset among the three. As shown in Table \ref{tab:TSF_main}, $\ell_1$ achieves the best results for 24- and 48-hour forecasting windows in ETTh1. While DTF-net also outperforms the single forecasting model, the linearity characteristic of $\ell_1$ is better suited for short-term predictions within ETTh1. In contrast, for long-term predictions, we demonstrate that DTF-net performs the best. In the case of Illness with a small dataset size, DTF-net also performs well without overfitting.

\subsubsection{Ablation Study}
\label{sec4.2.3}
\textbf{How to prevent RL overfitting? }
To mitigate the risk of overfitting in RL-based trend filtering, we introduce a reward sampling method. As shown in Figure \ref{fig:overfitting}, we observe that reward sampling prevents overfitting, achieving optimal performance with a reward sampling (Appendix \ref{appendix.b-4}).

\textbf{Empirical analysis on extreme value }
DNNs often generate smooth and averaged predictions as they typically optimize forecasting performance through empirical risk minimization. However, by incorporating accurately captured abrupt changes as additional information into the model, predictions are enhanced while representing both upward and downward signals instead of providing solely smooth estimates. To qualitatively assess the impact of abrupt changes on forecasting tasks, we compare two different trends: the original trends from DTF-net (red) and a version where 10\% of extreme values are excluded (blue). As shown in Figure \ref{fig:extreme value study}, this comparison demonstrates how incorporating abrupt changes can enrich forecasting by providing more detailed and accurate predictions.

\section{Conclusion}
We propose DTF-net, a novel RL-based trend filtering method directly identifying trend points. Traditional trend filtering methods struggle to capture abrupt changes due to their inherent approximations. To address this, we formalize the Trend Point Detection problem as an MDP and utilize RL within a discrete action space. The reward function is defined as the sum-of-squares loss from forecasting tasks inspired by reward function learning within the Gaussian distribution, allowing for the capturing of temporal dependencies around DTPs. DTF-net also tackles overfitting issues through random sampling. Compared to other trend filtering methods, DTF-net excels in identifying abrupt changes. In forecasting tasks, DTF-net enhances predictive performance without compromising the prediction output to be smooth.

\section*{Acknowledgments}
This work was partly supported by Institute of Information \& Communications Technology Planning \& Evaluation (IITP) grant funded by the Korea government (MSIT) (No. 2022-0-00984, Development of Artificial Intelligence Technology for Personalized Plug-and-Play Explanation and Verification of Explanation; No. 2022-0-00184, Development and Study of AI Technologies to Inexpensively Conform to Evolving Policy on Ethics; No. 2021-0-02068, Artificial Intelligence Graduate School Program (KAIST)).

\bibliographystyle{named}
\bibliography{ijcai24}

\clearpage
\appendix

\section{Extended Related Work}
\label{sec2}
Abrupt changes represent a shift in the slope of the trend and are categorized as extreme values. Focusing on this characteristic, we delve into algorithms specifically designed to identify points that share similar concepts with abrupt changes, such as structural breakpoints, change points, and anomaly points. Additionally, we discuss why these algorithms may not be suitable for trend-filtering tasks.

\subsection{Structural Break Point and Change Point Detection}
\subsubsection{Statistical Test}
A Structural Break Point (SBP) signifies an unexpected change in time series data. Various statistical algorithms can validate identified SBPs, including Chow \cite{chow1960tests}, Augmented Dickey-Fuller (ADF) \cite{dickey1979distribution}, and CUSUM \cite{ploberger1992cusum}. These methods confirm whether the detected points are structural breaks, relying on statistical tests using the F-value. In practical applications, algorithms should not only validate detected SBPs but also predict them in an online manner. The online approach involves predicting based on past to current state as data arrives incrementally over time, while the offline approach is prediction based on the entire dataset at once. Figure \ref{fig:mesh3}-(a) illustrates the results of detected SBPs in an offline manner based on the maximum F-value, comparing two sub-sequences while moving time points one by one. Each algorithm detects SBPs based on its own rule, resulting in irregular breakpoints that are challenging to adapt to the trend. This offline approach is highly sensitive to the input length and requires the heuristic definition of an additional threshold for breakpoint detection.

\subsubsection{Change Point Detection}
A change point denotes a specific time point or position within a time series where a significant shift or change occurs in the statistical properties of the data. Change Point Detection (CPD) algorithms specialize in identifying these shifts in distribution. One notable CPD approach is Bayesian Online Change Point Detection (BOCPD) \cite{adams2007bayesian}. This method detects points by maintaining a probability distribution over possible change points and updating this distribution as new data points are observed. However, BOCPD faces practical limitations due to its assumption of independence and the presence of temporal correlations between samples. Additionally, sensitivity to hyperparameters is a drawback of BOCPD \cite{han2019confirmatory}. Essentially, the probabilistic approach tends to overlook extreme values as outliers, leading to the detection of only the mean shift.

The Kolmogorov-Smirnov (KS) test-based segmentation \cite{gong2012kolmdogorov} assesses whether two data samples originate from the same underlying cumulative distribution. However, the KS algorithm is designed for offline detection, implying that it evaluates the entire time series at once. RED-SDS \cite{ansari2021deep} is a supervised learning method for segmentation, utilizing switching behavior to understand the underlying dynamical system in time series. However, as a supervised approach, RED-SDS requires a preceding labeling process to accurately train the model on correct change points. These segmentation-based CPD algorithms may miss important change points within a segment.

The clustering-based TICC algorithm \cite{hallac2017toeplitz} generates a sparse Gaussian inverse covariance matrix in clustered segments. However, TICC requires defining the number of clustering functions as a hyperparameter, necessitating prior knowledge and limiting the algorithm to detecting only a fixed number of change points. The Gaussian Process (GP) and Bayesian-based ADAGA algorithm \cite{adaga} define both the mean and covariance structure of the temporal process based on the GP kernel. However, ADAGA is sensitive to the GP-kernel type, which also requires prior knowledge. These GP-based CPD algorithms have limitations in capturing non-smoothness, as GP approximates smoothness using mean-square differentiable functions \cite{banerjee2003smoothness}.

\begin{figure}
    \centering
    \subfloat[Statistic-based algorithms]{\includegraphics[clip, trim=0 0 0 0.9cm, width=0.45\textwidth]{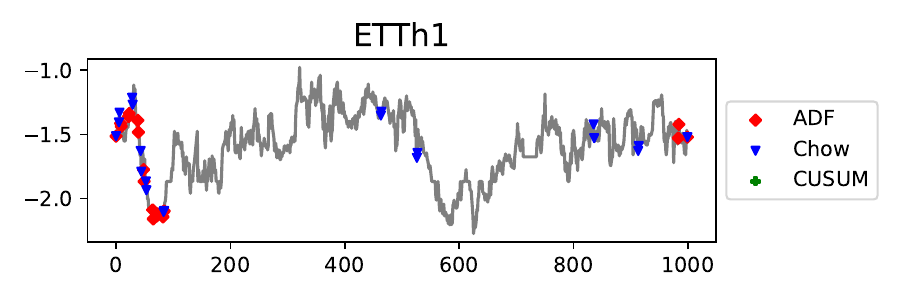}}\\
    \subfloat[CPD algorithms]{\includegraphics[clip, trim=0 0 0 0.9cm, width=0.45\textwidth]{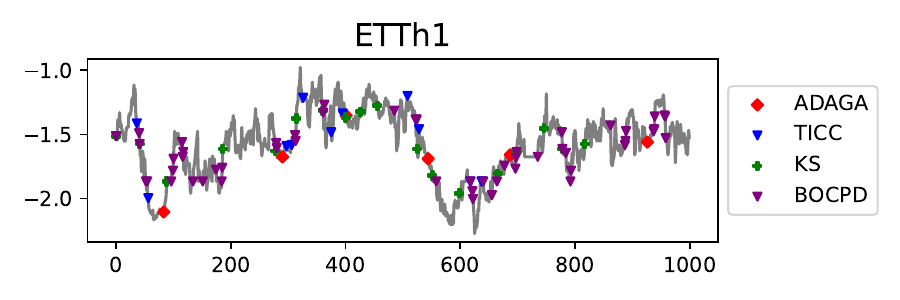}}\\
    \subfloat[Limitation of CPD on Trend Filtering]{\includegraphics[clip, trim=-0.5cm -0.25cm -0.5cm 0.9cm, width=0.45\textwidth]{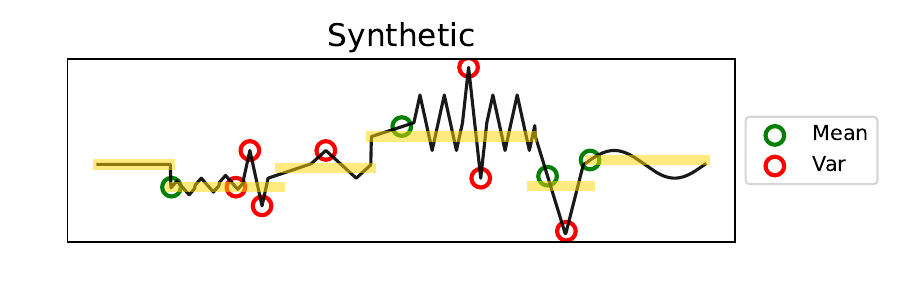}}
    \caption{\textbf{Limitation of SBP and CPD algorithms in trend filtering.} Figures (a) and (b) illustrate how statistical-based and CPD algorithms detect abrupt changes based on their own criteria in a real-world dataset, ETTh1. The detected points appear irregular, making them challenging to adapt to the trends. In general, Figure (c) illustrates the difficulties CPD algorithms face when identifying points where there is no mean shift and only variance perturbation occurs, as highlighted by the red circles.}
    \label{fig:mesh3}
\end{figure}

\begin{figure*}
    \centering
    \subfloat[DTF-net]{\includegraphics[width=0.33\textwidth]{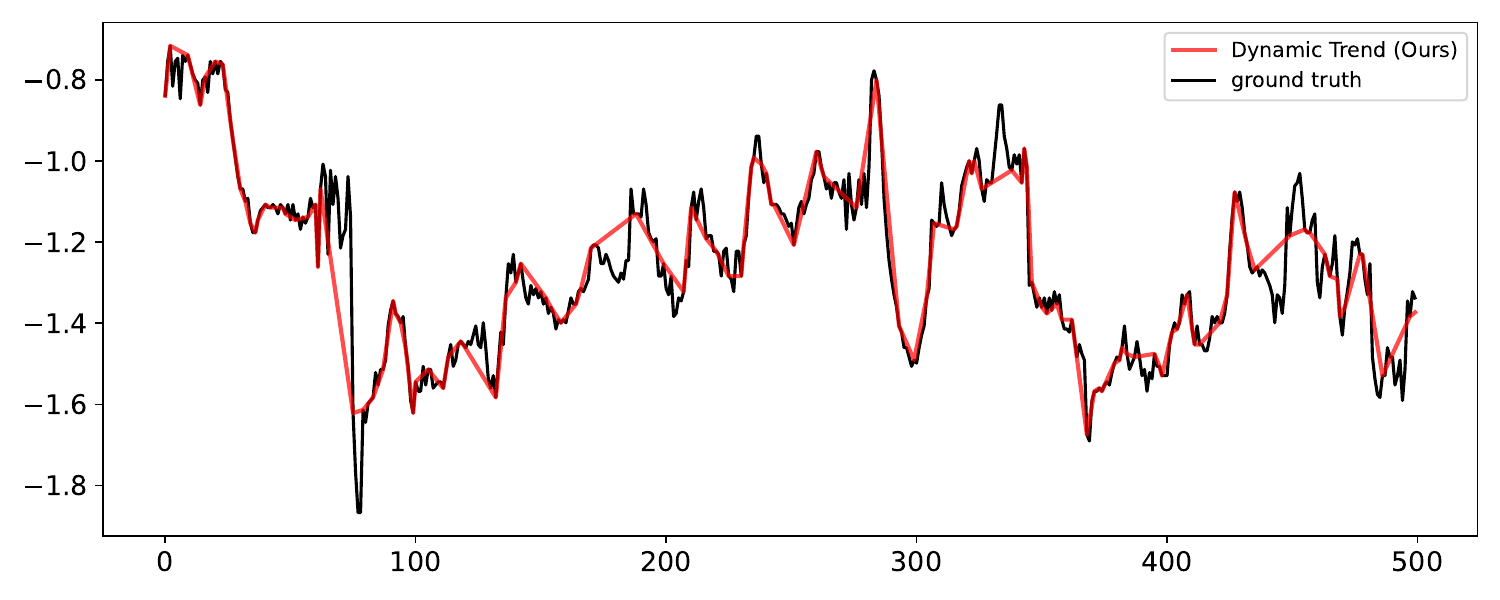}}
    \subfloat[ADAGA]{\includegraphics[width=0.33\textwidth]{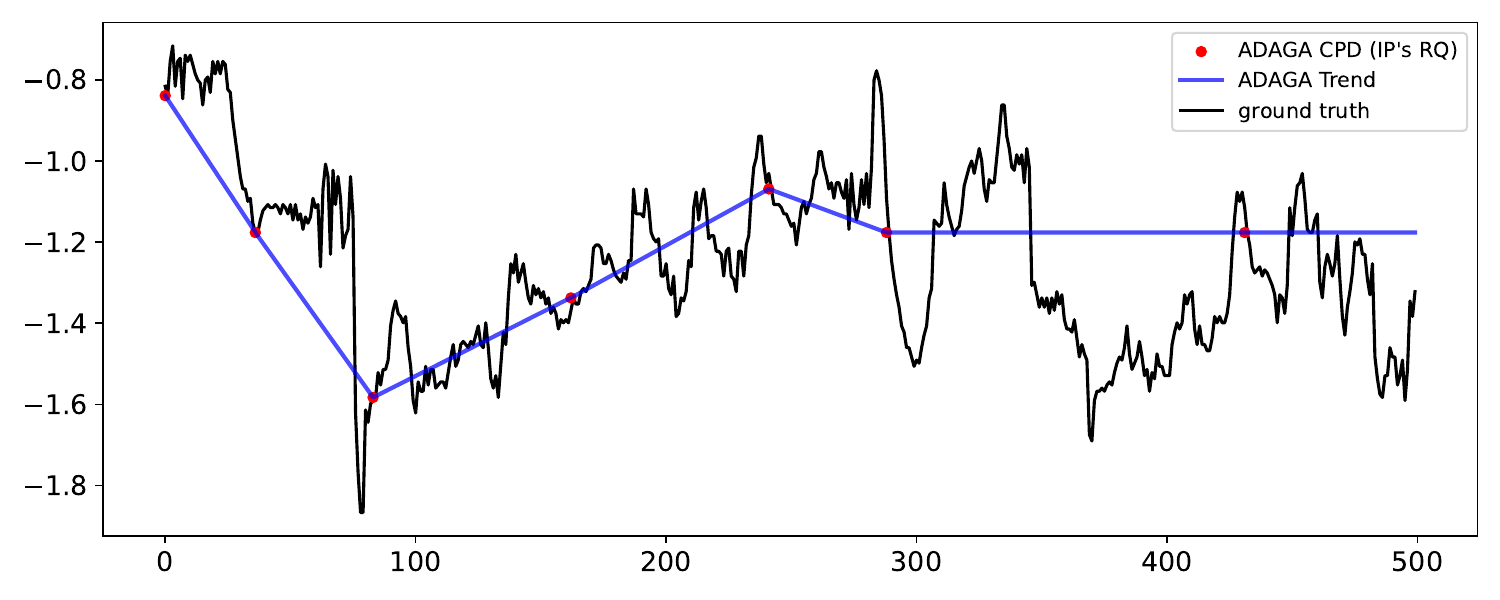}}
    \subfloat[BOCPD]{\includegraphics[width=0.33\textwidth]{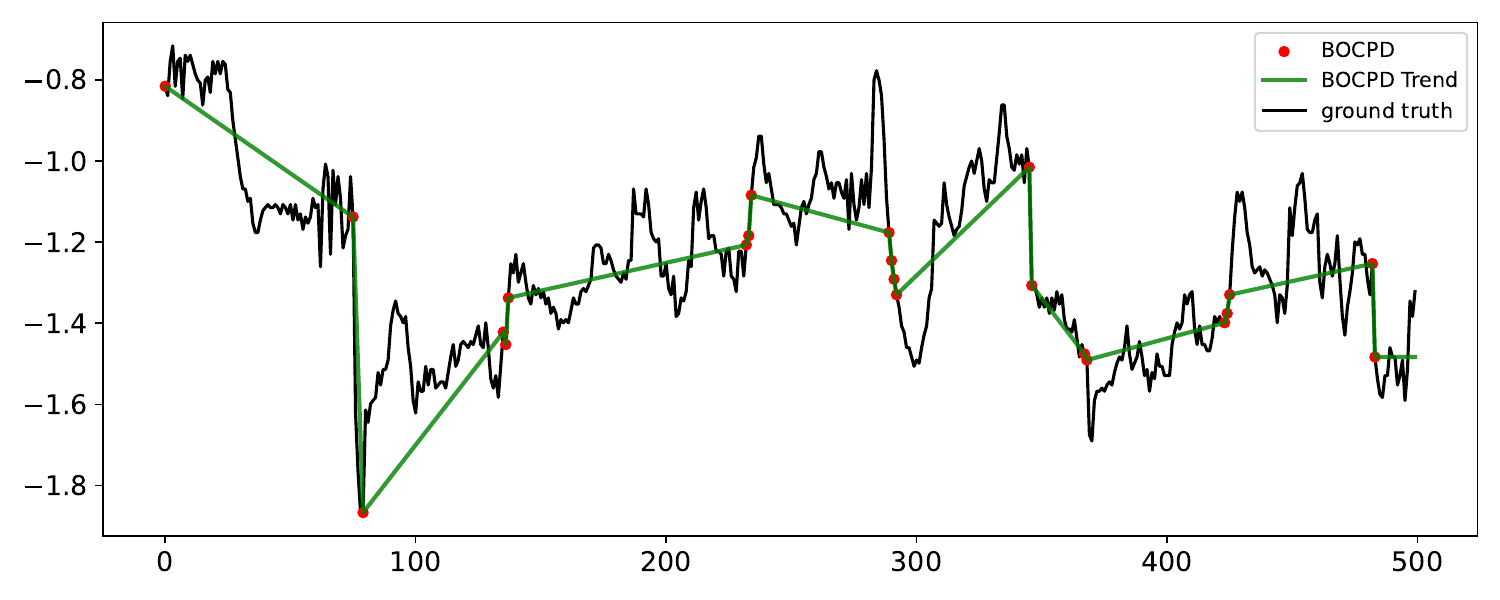}}\\
    \subfloat[TimesNet]{\includegraphics[width=0.33\textwidth]{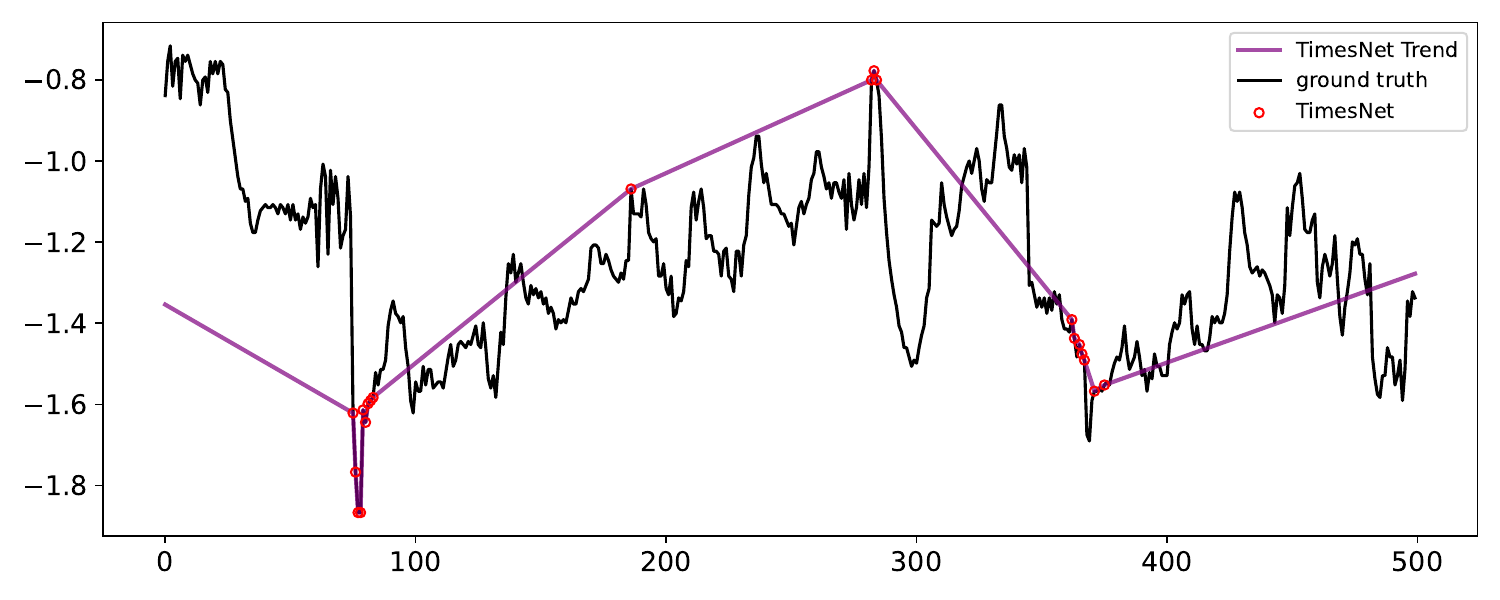}}
    \subfloat[Anomaly Transformer]{\includegraphics[width=0.33\textwidth]{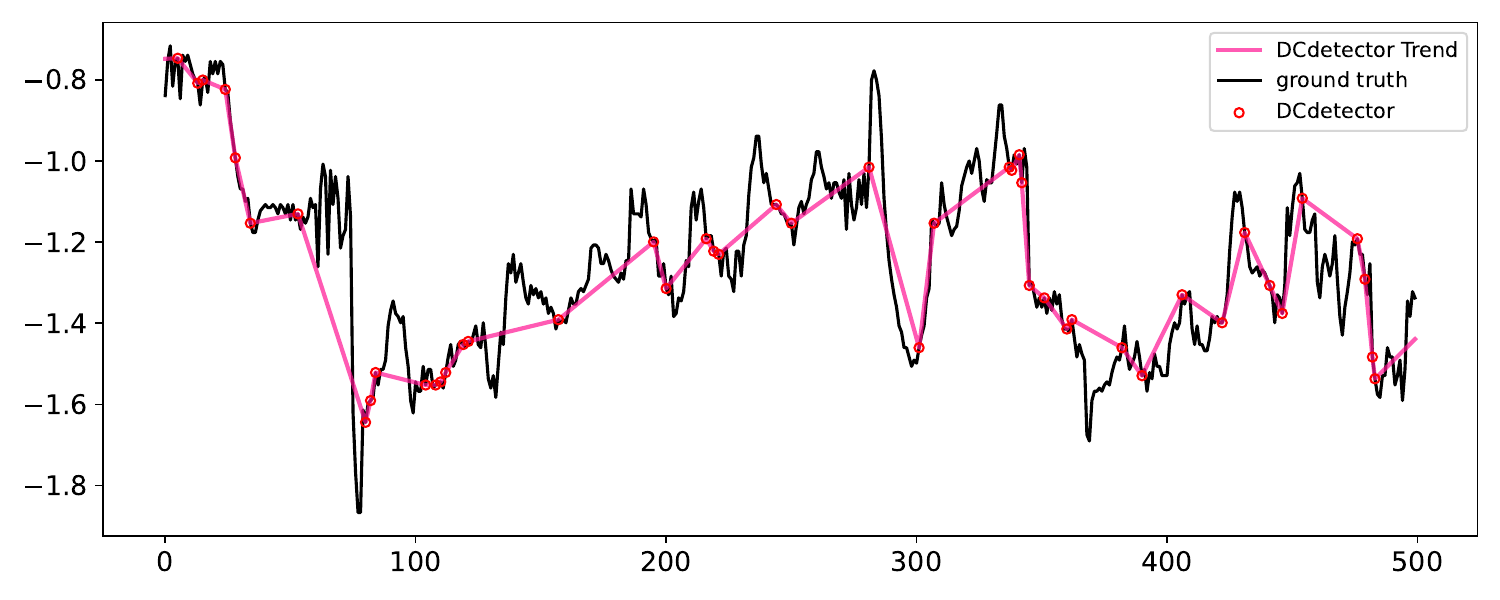}}
    \subfloat[DCdetector]{\includegraphics[width=0.33\textwidth]{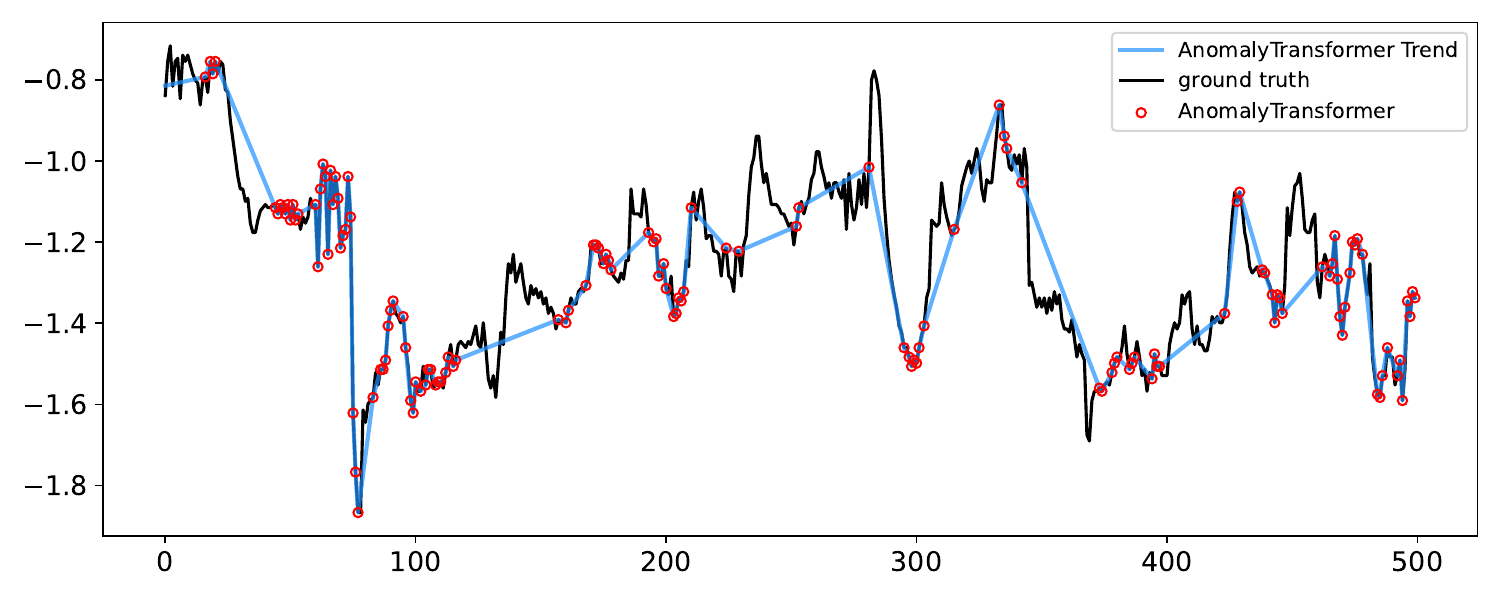}}
    \caption{\textbf{Limitation of CPD and anomaly detection algorithms in trend filtering.} The figure presents comparative experiments involving DTF-net, CPD, and Anomaly Detection for trend filtering on the ETTh1 dataset. CPD algorithms excel in capturing change points related to mean shifts but face challenges when dealing with points exhibiting variance shifts. While anomaly detection methods prioritize extreme values over CPD, they focus solely on abnormal points and lack the ability to detect midpoints that are crucial for reflecting trends. In contrast, DTF-net dynamically performs trend filtering by incorporating abrupt changes, as exemplified around the 300th point.}
    \label{fig:cpd}
\end{figure*}

Empirically, Figure \ref{fig:mesh3}-(b) highlights the detected change points specifically in the mean shift on the ETTh1 dataset across different CPD algorithms. In general, as depicted in Figure \ref{fig:mesh3}-(c), probabilistic-based CPD algorithms exhibit a notable limitation by primarily detecting mid-points (green) in mean shifts rather than the vertices of abrupt changes (red), which are solely due to variance perturbations. This limitation leads to an over-smoothing issue in trend extraction, as further illustrated in Figure \ref{fig:cpd}-(b) and -(c).

\subsection{Anomaly Detection}
Anomaly detection involves identifying abnormal points or sub-sequences distinguished from the regular patterns found in the majority of the dataset. As the goal of capturing extreme values aligns with DTF-net's objectives, we conduct a comparison with advanced anomaly detection algorithms, including TimesNet \cite{wu2023timesnet}, Anomaly Transformer \cite{xu2022anomaly}, and DCdetector \cite{yang2023dcdetector}.

TimesNet \cite{wu2023timesnet} utilizes a Temporal Variation Modeling methodology, which is a transforming technique of the original 1D time series into a 2D space to capture multi-periodicity and complex interactions in time series data. However, due to its focus on complex patterns, TimesNet tends to predict extreme anomaly points and might not perform optimally in trend filtering tasks, as illustrated in Figure \ref{fig:cpd}-(d). Anomaly Transformer \cite{xu2022anomaly} introduces a novel Anomaly-Attention mechanism, employing a Gaussian kernel and minimax strategy to enhance normal-abnormal distinguishability. Nevertheless, due to the Gaussian kernel's smoothness, it fails to detect some important peaks at 80, 300, and 350. DCdetector \cite{yang2023dcdetector} adopts a unique dual attention asymmetric design for contrastive learning, creating a permutated environment through patching techniques. Leveraging contrastive learning, the DCdetector exhibits a more generalized architecture compared to the Anomaly Transformer, but it still faces challenges in detecting peaks around 300. Additionally, the trend filtering task faces challenges in smoothing noisy sub-sequences, particularly around axes 70 and 450. All the mentioned anomaly detection algorithms exhibit sensitivity to thresholds, and even with an appropriate threshold, points are often missed in peak regions, such as around 300. In summary, while anomaly detection algorithms often miss crucial mid-points of changes, CPD algorithms may overlook essential change points in variance shifts.

\begin{figure*}[h]
    \captionsetup[subfigure]{oneside,margin={0cm,0cm}}
    \subfloat[DQN]{\includegraphics[clip, trim= 0 0 5.5cm 0, width=0.3\textwidth]{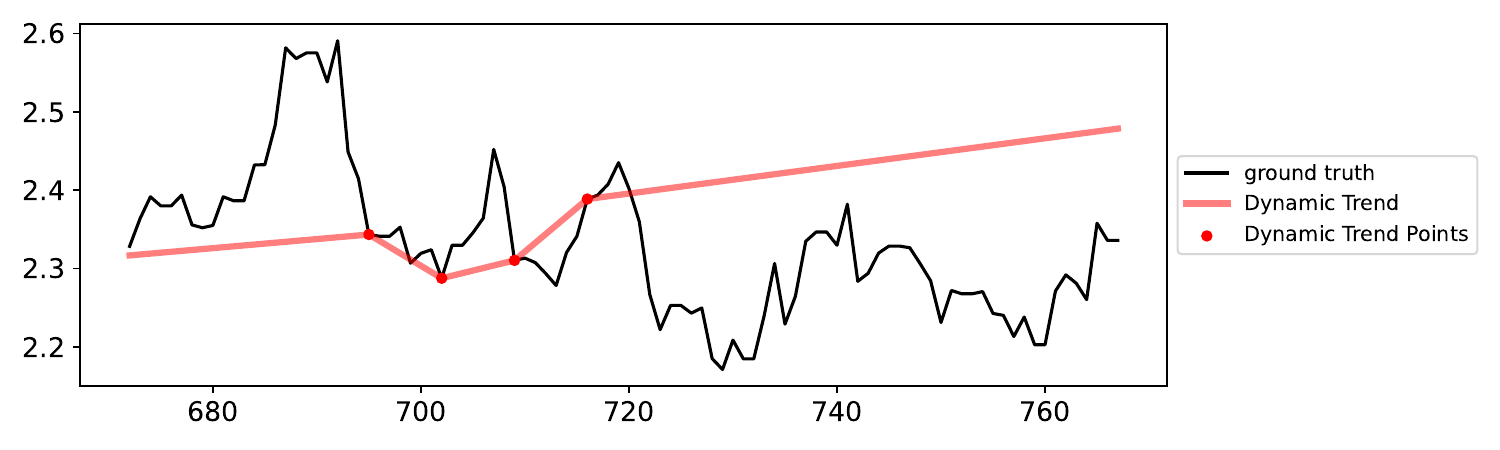}}
    \subfloat[A2C]{\includegraphics[clip, trim= 0 0 5.5cm 0, width=0.3\textwidth]{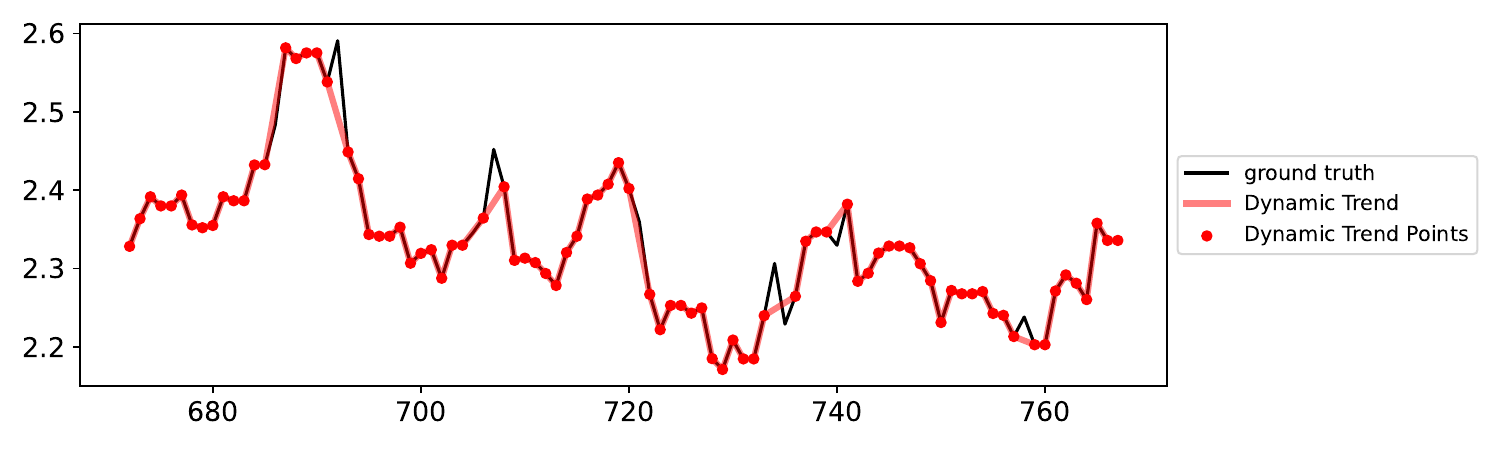}}
    \captionsetup[subfigure]{oneside,margin={0cm,1cm}}
    \subfloat[PPO]{\includegraphics[width=0.38\textwidth]{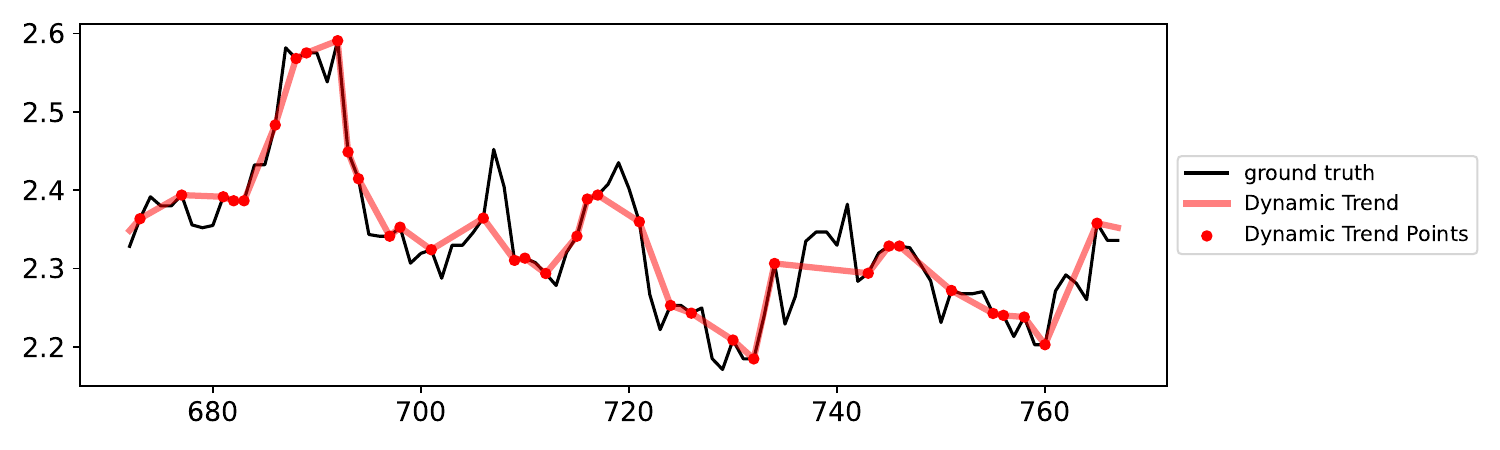}}
    
    \caption{\textbf{Dynamic trend filtering with different RL algorithms.} The figure illustrates different trend-filtering outcomes obtained using various RL algorithms. DQN demonstrates an underfitting trend, while A2C exhibits an overfitting trend.}
    \label{fig:trend_length}
\end{figure*}

\section{Extended Related Work on RL}
\label{app: RL related work}

\subsection{RL for Discrete Action Space}
In Reinforcement Learning (RL), the agent consistently interprets and responds to an environment to maximize a specified reward. The environment is a series of events, prompting the agent to select actions that it deems most conducive to achieving the maximum long-term reward. Within the framework of Markov Decision Processes (MDP), the agent learns optimal actions to enhance the cumulative reward associated with each state. A Markov chain illustrates the relationships among sequences of events, providing crucial information for models designed specifically for time series analysis, where discerning underlying trends is imperative \cite{mcmc,sutton2018reinforcement,liu2022finrl_meta}.

The probability distribution that an action is selected for each state is referred to as the policy, denoted as $\pi(A|S) = \operatorname{Pr}(A|S)$.
The state-value function, denoted as $v_{\pi}(S)$, represents the expected return value following policy $\pi$ from state $S$ and is defined as follows,
\begin{equation}
    \begin{aligned}
        v_{\pi}(S) & =\mathrm{E}_{\pi}[G|S]=\mathrm{E}_{\pi}[\mathcal{R}'+\gamma v_{\pi}(S')|S] \\
                   & =\sum_{A} \pi(A|S) (\mathcal{R}+\gamma \sum_{S'} \operatorname{Pr}(S'|S, A) v_{\pi}(S')),
\end{aligned}
\end{equation}
where $S'$ and $R'$ represent the next state and reward, and $G=\sum_{k=0}^{\infty} \gamma^k \mathcal{R}'_{k}$.
The action-value function, denoted as $q_{\pi}(S,A)$, is the expected value of the return from taking action $A$ in state $S$:
\begin{equation}
    \begin{aligned}
        q_{\pi}(S, A) & =\mathrm{E}_{\pi}[G|S, A]\\
                      & =\mathrm{E}_{\pi}[\mathcal{R}'+\gamma q_\pi(S', A')|S] \\
                      & =\mathcal{R}+\gamma \sum_{S} \operatorname{Pr}(S'|S, A) \sum_{A'} \pi(A'|S') q_\pi(A'|S'),
\end{aligned}
\end{equation}
based on the Bellman equation,
\begin{equation}
    v_\pi(S) = \sum_A \pi(S,A) \sum_{S'} \operatorname{Pr}(S' | S,A) ( \mathcal{R} + \gamma v_\pi (S')).
    \label{eq:bellman}
\end{equation}

In model-based RL, the agent constructs an internal model or representation of the environment, requiring state transition probabilities and corresponding rewards for each state. Model-based methods involve dynamic programming or Monte Carlo simulation for planning based on the acquired model. In contrast, model-free RL teaches agents to interact directly with the environment, mapping states to actions without estimating transition dynamics. Model-free methods rely on trial-and-error learning, where the agent explores the environment, receives feedback in the form of rewards, and adjusts its strategy over time based on observed outcomes. Common model-free approaches include Deep Q Networks (DQN) and Policy Gradient methods. In this research, our focus is on model-free approaches.

\textbf{DQN } To address the challenges of applying Q-learning to environments with high-dimensional state spaces, DQN combines Q-learning with DNNs. It uses DNNs to approximate the Q-function, which represents the expected cumulative future rewards for taking an action in a given state. With two separate networks, the Q-network and the target network, DQN stabilizes the learning process by providing fixed targets for Q-values during temporal difference updates. The Q-network is used to select actions, while the target network is periodically updated with the Q-network's weights. Additionally, with a defined replay buffer, DQN stores past experiences, such as the state, action, reward, and next state of the agent. During training, random batches of experiences are sampled from the replay buffer, breaking the temporal correlation between consecutive samples and improving data efficiency \cite{mnih2015human}.

\textbf{A2C } To overcome the limitation of increasing variance as the episode prolongs, Actor-Critic methods do not use a replay buffer but instead learn actions directly. The Actor-Critic framework defines both an actor and a critic, with the policy network estimating actions and the value function estimating state-action values, respectively. A2C is one such Actor-Critic algorithm that combines elements of both policy iteration (the actor) and value iteration (the critic) to improve stability and sample efficiency. The actor is responsible for selecting actions based on the current policy, while the critic evaluates the value of state-action pairs. Using policy gradients, A2C updates the actor's policy to maximize the expected cumulative reward. For updating value estimates, A2C employs temporal difference errors, which represent the difference between the observed return and the current estimate of the value function \cite{A2C}.

\textbf{PPO }
Another actor-critic method, PPO, is designed to optimize policies for environments with both continuous and discrete action spaces. It employs a surrogate objective function that combines policy improvement with a clip ratio, controlling the extent of policy updates in each iteration to enhance stability. The clip ratio acts as a trust region constraint, preventing large policy deviations. PPO utilizes policy gradient updates and multiple epochs for training, involving the collection of experiences, computation of policy updates, and application to the policy. Additionally, it often incorporates value function optimization to better estimate state values. Known for its stability and robustness, PPO has been widely applied to diverse tasks, demonstrating effectiveness in both continuous and discrete action environments \cite{schulman2017proximal}.

\subsection{Motivation of Using RL for Trend Filtering}
Stock trading represents one of the most representative problems modeled within an MDP using time series data \cite{liu2022finrl_meta}. The decision-making process adheres to the Markov property, where actions depend solely on the immediate preceding state and are not influenced by the past. The optimal trading points are equivalent to abrupt changes and extreme values: buying at the lowest turning point and selling at the highest to maximize profit. To solve this MDP-modeled stock trading problem, RL stands out as one of the most popular methods capable of directly detecting trading points through an agent's action prediction. Motivated by these equivalences, we formulate the Trend Point Detection problem as an MDP and use RL to solve it, aiming to identify essential points for trend filtering.

%\section{Discussion on DTF-net}
\subsection{Gaussian Distribution and Reward Function} 
According to \cite{nips_gaussian_reward}, in a continuous state space, the Bellman Equation (\ref{eq:bellman}) can be generalized by substituting sums with integral when the policy is deterministic:
\begin{equation}
    \begin{split}
        v_\pi(S)  & = \int \operatorname{Pr}(S'|S, \pi(S)) (\mathcal{R}_{\pi(S)} + \gamma v_\pi(S')) dS'\\
        & = \int \operatorname{Pr}(S'|S, \pi(S)) \mathcal{R}_{\pi(S)} dS' \\ 
        &+ \gamma \int \operatorname{Pr}(S'|S, \pi(S)) v_\pi(S') dS',
    \end{split}
    \label{integral_bellman}
\end{equation}
where $\pi(S)$ is consecutive states $S'$ when following the policy $\pi$. Based on Equations (\ref{mse}), (\ref{gp}), the reward is a Gaussian distribution over consecutive state variable $S'$, where $R_{\pi(S)}$ follows $\mathcal{N}(y_t, \tau^2)$.  

Then, policy update is defined as follows,
\begin{equation}
    \pi(S) \leftarrow argmax_{A} \int \operatorname{Pr}(S'|S,A) (R + \gamma v(S')) dS'.
\end{equation}
Therefore, the policy is optimized based on the distribution of time series data, following a Gaussian distribution. This allows the agent to predict actions based on time series patterns, enabling different trend point detections in distinct sub-sequences while capturing temporal dependencies.

\begin{figure}
    {\includegraphics[width=0.49\textwidth, clip, trim= 0cm 6cm 0 0cm]{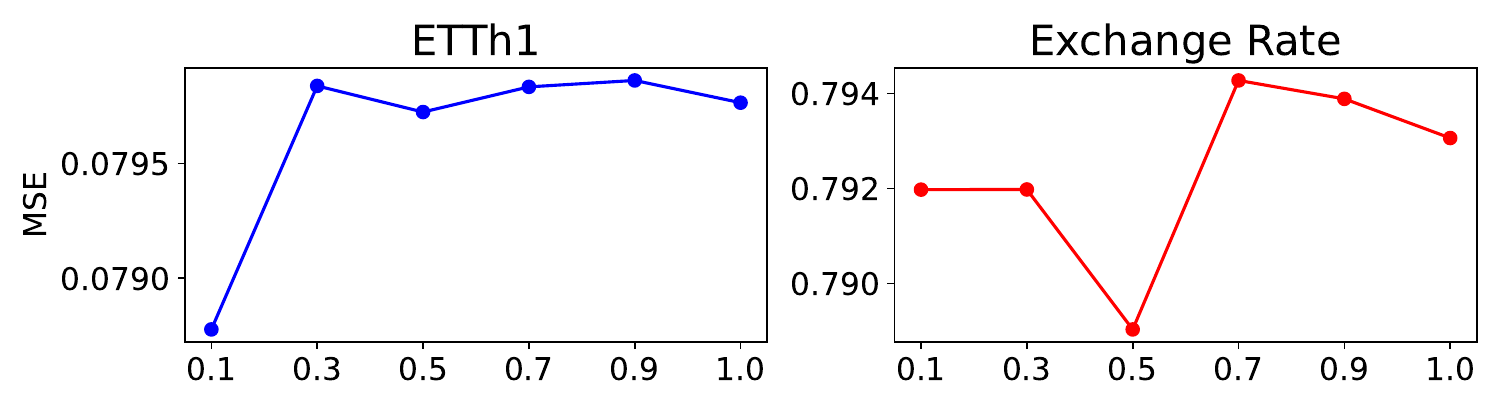}}
    
    \caption{\textbf{DTF-net prevents overfitting problems.} The figure illustrates each dataset has an optimal reward sampling ratio while addressing the overfitting issue. The $x$-axis denotes the reward sampling ratio, while the $y$-axis represents the MSE.}
    \label{fig:overfitting}
\end{figure}

\subsection{Sampling and Penalty Reward}
\label{appendix.b-4}
In RL, positive rewards and penalty rewards are utilized to shape the behavior of the agent by providing feedback on its actions. Positive rewards are assigned to actions that bring the agent closer to the desired goal or enhance its performance, encouraging the agent to repeat those actions. Conversely, penalty rewards are associated with actions that should be avoided or are suboptimal. These penalty rewards discourage the agent from taking undesirable actions, guiding it away from unfavorable behavior. In DTF-net, a penalty reward in the form of a negative sum-of-squares function is employed. However, penalizing at every timestep may cause the agent to become overly fitted to the original time series data, as it aims to minimize the penalty consistently. To address this limitation, a sampling method is employed to determine the penalizing interval, providing a solution to overcome this challenge. We empirically demonstrate that the optimal reward sampling ratio enhances the performance of DTF-net, as shown in Figure \ref{fig:overfitting}.

\subsection{Computational complexity }
While RL for trend filtering enhances dynamic trend extraction and abrupt change capture, it comes with increased computational complexity, rising from $\mathcal{O}(1)$ to $\mathcal{O}(n)$ compared to other trend filtering methods. This increase is due to the batch learning of deep networks in RL, while other methods optimize the entire sequence with a single cost function at once. However, we address this computational inefficiency through three strategies. First, we employ an MLP policy network, which has lower computational costs compared to convolutional or recurrent-based architectures while enhancing trend-filtering performance. Second, we incorporate random sampling in DTF-net for the state and reward, utilizing only part of the data and not calculating the reward in every step. Third, we keep the episode iteration at less than 10, mitigating the overfitting issue and reducing time complexity simultaneously. Additionally, from the perspective of the data-intensive nature of deep networks, DTF-net demonstrates robust performance with a small number of datasets, such as synthetic or illness data, using a simple MLP policy network. In summary, DTF-net effectively minimizes the increase in computational complexity when applying RL for trend filtering.

\onecolumn
\section{Extended Experiments on TSF}
\label{app: TSF}
\subsection{TSF Dataset}
\label{app: TSF dataset}
Based on the ADF test, we specifically choose non-stationary datasets in the TSF benchmark to evaluate DTF-net's ability to capture abrupt changes in Table \ref{tab:TSF_main} \cite{liu2022nonstationary}.

\begin{table}[ht]
    \centering
    \begin{tabular}{lccc}
    \hline
        Dataset & Variable Number & Sampling Frequency & ADF Test Statistic \\
        \hline
         Exchange & 8 & 1Day & -1.889\\
         ILI & 7 & 1Week & -5.406\\
         ETT & 7 & 1Hour / 15Minutes & -6.225\\
         Electricity & 321 & 1Hour & -8.483\\
         Traffic & 862 & 1Hour & -15.046\\
         Weather & 21 & 10Minutes & -26.661\\
         \hline
    \end{tabular}
    \caption{\textbf{Summary of TSF datasets.} A smaller ADF test statistic indicates a more stationary dataset.}
    \label{tab:my_label}
\end{table}

\subsubsection{ADF Test}
The Augmented Dickey-Fuller (ADF) test is a statistical test used to determine whether a unit root is present in a univariate time series dataset. The presence of a unit root indicates that the time series data has a stochastic trend and is non-stationary. The null hypothesis of the ADF test is that a unit root is present, suggesting non-stationarity. If the test statistic is significantly below a critical value, the null hypothesis is rejected, indicating that the time series is stationary after differencing.

The ADF test is conducted using the following regression equation:
$$\Delta y_t = \alpha + \beta t + \gamma y_{t-1} + \delta_1\Delta y_{t-1} + ... + \delta_{p-1}\Delta y_{t-p+1} + \epsilon_t,$$
where $\Delta y_t$ is the differenced series at time $t$, $y_{t-p}$ is lagged value, $t$ is a time trend variable, $\delta_i$ are coefficients on lagged differences terms,  and $\epsilon_t$ is residual term.

\subsubsection{Benchmark Dataset}
\begin{itemize}
     \item \texttt{Exchange} dataset comprises daily exchange rates of eight foreign countries, spanning from 1990 to 2016 \cite{zhou2021informer}.
    \item \texttt{Illness} dataset includes patient data for influenza illness recorded weekly from the US Centers for Disease Control and Prevention between 2002 and 2021. It represents the ratio of patients seen with influenza-like illness to the total number of patients \cite{wu2021autoformer}.
    \item \texttt{ETT} (Electricity Transformer Temperature) dataset consists of recorded over a one-hour/minute period. Each data point includes the target value `Oil Temperature' along with six power load features \cite{zhou2021informer}.
    \item \texttt{Electricity} dataset consumption data, measured in kilowatt-hours (Kwh), was gathered from 312 clients. To account for missing values, the data was converted into hourly consumption over a span of two years \cite{zhou2021informer}.
    \item \texttt{Traffic} dataset comprises hourly information from the California Department of Transportation, detailing road occupancy rates measured by various sensors on San Francisco Bay Area freeways \cite{wu2021autoformer}.
    \item \texttt{Weather} dataset includes local climatological data for nearly 1,600 U.S. locations, with data points collected every hour over a span of four years \cite{zhou2021informer}.
\end{itemize}

\newpage
\subsection{TSF Baselines.}
\subsubsection{TSF Baseline Models}
\begin{itemize}
    \item \texttt{Autoformer}, a Transformer-based method, utilizes a decomposition and auto-correlation mechanism based on fast Fourier transform to learn the temporal patterns of time series data \cite{wu2021autoformer}.
    \item \texttt{FEDformer}, a Transformer-based method, introduces a Mixture of Experts (MOE) for seasonal-trend decomposition and utilizes frequency-enhanced block/attention with Fourier and Wavelet Transform \cite{zhou2022fedformer}.
    \item \texttt{DLinear}, relying solely on linear layers, decomposes the original input into trend and remainder components using global average pooling. Two linear layers are applied to each component, and the resulting features are summed up to obtain the final prediction \cite{zeng2022transformers}.
    \item \texttt{NLinear} employs a straightforward normalization technique to tackle the train-test distribution shift in the dataset. This technique involves subtracting the last value from the input and adding it back before making the final prediction, all while using only a single linear layer \cite{zeng2022transformers}.
    \item \texttt{PatchTST}, a transformer-based method, comprises two components: the segmentation of time series into subseries-level patches and a channel-independent structure. PatchTST can capture local semantic information and benefits from longer look-back windows, utilizing patching techniques \cite{nie2022time}.
\end{itemize}

\subsubsection{Hyper-parameters}
\begin{table*}[ht]
    \centering
    \begin{adjustbox}{width=1\textwidth}
    \begin{tabular}{c|c|c|c|c|c|c|c|c}
        \hline
        Dataset & Input Length & Prediction Length & Forecasting Model & Reward Ratio & Learning Rate & RL Steps & Forecasting Epoch & Max Sequence Length \\
        \hline
        \multirow{6}{*}{Exchange Rate} & 336 & 720 & DLinear & 0.4 & 5e-4 & 10000 & 15 & 3000 \\
        & 336 & 336 & DLinear & 0.4 & 5e-4 & 10000 & 15 & 3000 \\
        & 336 & 192 & DLinear & 0.4 & 1e-4 & 10000 & 15 & 3000 \\
        & 336 & 96 & NLinear & 0.4 & 1e-4 & 3000 & 15 & 3000 \\
        & 336 & 48 & NLinear & 0.4 & 1e-4 & 3000 & 15 & 3000  \\
        & 336 & 24 & NLinear & 0.4 & 1e-4 & 3000 & 15 & 3000 \\
        \hline
        \multirow{6}{*}{ETTh1} & 336 & 720 & NLinear & 0.1 & 1e-3 & 10000 & 20 & 3000 \\
        & 336 & 336 & NLinear & 0.1 & 9e-4 & 10000 & 15 & 3000 \\
        & 336 & 192 & NLinear & 0.1 & 3e-4 & 10000 & 15 & 3000 \\
        & 336 & 96  & NLinear & 0.1 & 5e-4 & 1000  & 15 & 3000 \\
        & 336 & 48  & NLinear & 0.1 & 5e-4 & 1000  & 15 & 3000 \\
        & 336 & 24  & NLinear & 0.1 & 5e-4 & 1000  & 15 & 3000 \\
        \hline
        \multirow{3}{*}{Illness} & 104 & 60 & NLinear & 0.1 & 1e-3 & 1000  & 15 & 300 \\
        & 104 & 48 & NLinear & 0.1 & 1e-3 & 5000  & 15 & 300 \\
        & 104 & 24 & NLinear & 0.1 & 1e-3 & 10000 & 15 & 300 \\
        \hline
    \end{tabular}
    \end{adjustbox}
    \caption{\textbf{Hyper-Parameters.}}
    \label{tab:my_label}
\end{table*}

The experiments are conducted under the same conditions, employing the Adam optimizer, MSE loss function, 15 epochs, and batch sizes of 32. The input sequence length for linear models is set to 336, while Transformer models utilize an optimal length of 96. The learning rate for the baseline model follows the values outlined in Table \ref{tab:my_label}, and in cases of overfitting, it is adjusted to $1e-4$. The seed number is arbitrarily set to 2023.

\subsection{Dynamic Trend Extraction of DTF-net}
\begin{figure}[h]
    \subfloat[Exchange (pred=48)]{\includegraphics[clip, trim= 0 0 5.5cm 0, width=0.33\textwidth]{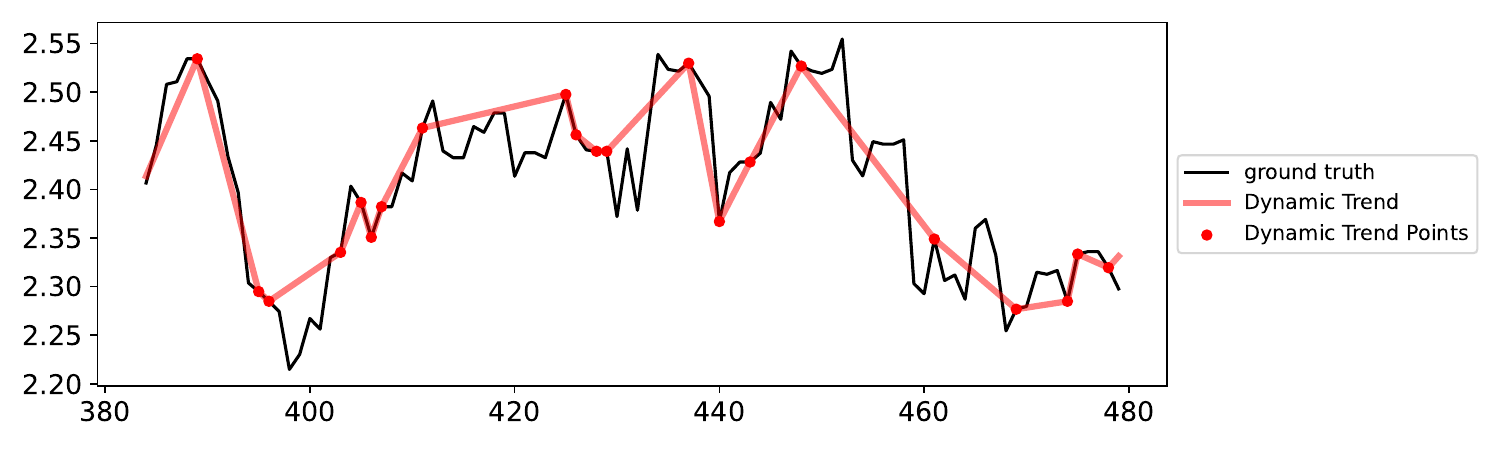}}
    \subfloat[Exchange (pred=192)]{\includegraphics[clip, trim= 0 0 5.5cm 0, width=0.33\textwidth]{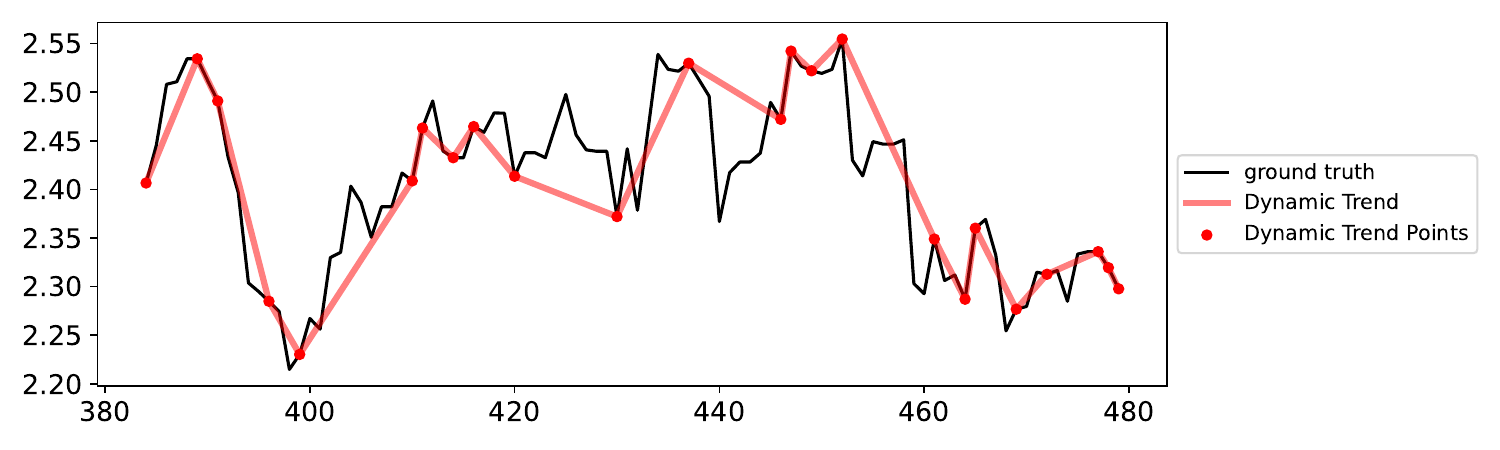}}
    \captionsetup[subfigure]{oneside,margin={0cm,1cm}}
    \subfloat[Exchange (pred=720)]{\includegraphics[clip, trim= 0 0 5.5cm 0, width=0.33\textwidth]{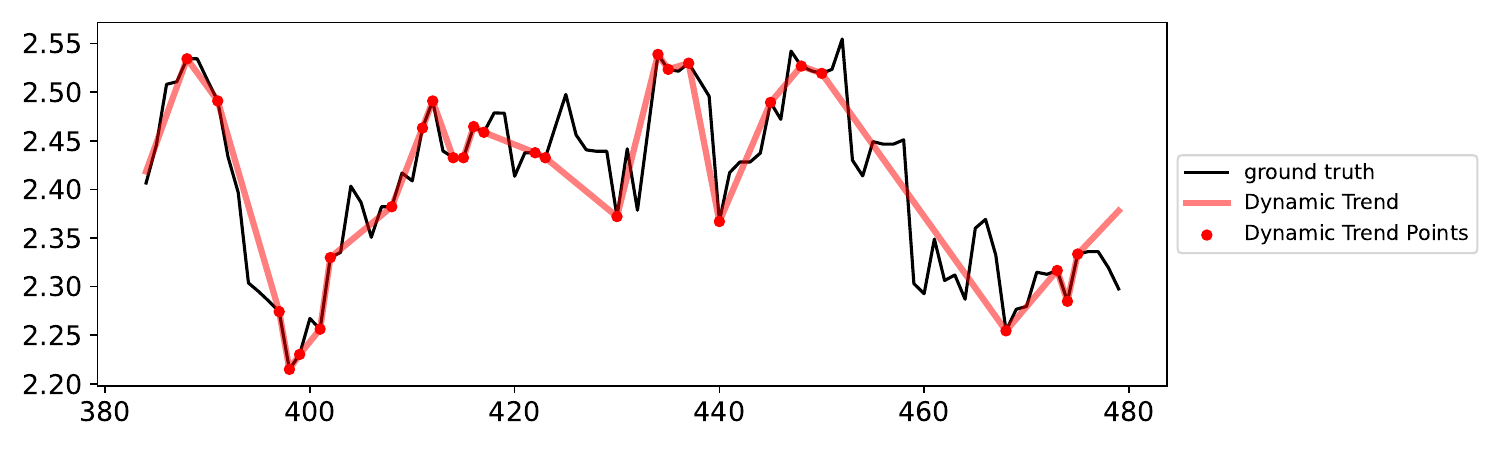}} 
    \caption{\textbf{Dynamic trend filtering of DTF-net with different prediction length.} The figure demonstrates that DTF-net extracts various trends based on different prediction horizon lengths on the Exchange Rate Dataset.}
    \label{fig:dynamic_trend}
\end{figure}
In comparison to traditional trend filtering methods, DTF-net extracts dynamic trends with varying levels of smoothness for each sub-sequence. This is achievable by adjusting the prediction length, as shown in Figure \ref{fig:dynamic_trend}.

\clearpage
\section{Extended Ablation Study on DTF-net}

\subsection{Ablation Study on State}
\label{app: state}

\begin{table*}[ht]
    \centering
    \begin{adjustbox}{width=0.9\textwidth}
    \begin{tabular}{c|c||cc||cc|cc|cc||cc}  
    \hline \multirow{2}{*}{State} & Episode & \multicolumn{2}{c||}{ non-sequential } & \multicolumn{2}{c|}{ non-sequential } & \multicolumn{2}{c|}{ sequential } & \multicolumn{2}{c||}{ sequential } & \multicolumn{2}{c}{\multirow{2}{*}{zero padding}} \\
    \cline{2-10}
    & length & \multicolumn{2}{c||}{dynamic}  & \multicolumn{2}{c|}{static} & \multicolumn{2}{c|}{dynamic} & \multicolumn{2}{c||}{static} & \\
    \hline \multicolumn{2}{c||}{Metric} & MSE & MAE & MSE & MAE & MSE & MAE & MSE & MAE & MSE & MAE \\
    \hline 

    \multirow{6}{*}{Exchange} & 24 & \bf 0.0250 & \bf 0.1198 & $0.0263$ & $0.1228$ & $0.0264$ & $0.1231$ & \underline{0.0263} & \underline{0.1227} & 0.0259 & 0.1215\\
    & 48 & \bf 0.0487 & \bf 0.1658 & $0.0500$ & $0.1697$ & \underline{0.0496} & \underline{0.1689} & $0.0501$ & $0.1699$ & \underline{\bf 0.0486} & \underline{\bf 0.1655}\\
    & 96 & \underline{0.0983} & \underline{0.2349} & $0.0995$ & $0.2363$ & $0.0994$ & $0.2363$ & \bf 0.0982 & \bf 0.2348 & 0.0983 & 0.2350\\
    & 192 & \bf 0.1983 & \bf 0.3583 & $0.1986$ & $0.3587$ & $0.2013$ & $0.3607$ & \underline{0.1984} & \underline{0.3582} & 0.2003 & 0.3598\\
    & 336 & \underline{0.3160} & \underline{0.4561} & $0.3163$ & $0.4562$ & $0.3166$ & $0.4562$ & \bf 0.3140 & \bf 0.4541 & 0.3166 & 0.4564\\
    & 720 & $ 0.7933 $ & $ 0.6874 $ & $0.7922$ & \textbf{0.6822} & \underline{0.7903} & $0.6878$ & \textbf{0.7893} & \underline{0.6874} & \underline{\bf 0.7889} & \underline{\bf 0.6871}\\
    \hline
    \multirow{6}{*}{ETTh1} & 24 & \bf 0.0253 & \bf 0.1205 & $0.0264$ & $0.1228$ & \underline{0.0255} & \underline{0.1205} & $0.0259$ & $0.1214$ & 0.0258 & 0.1216\\
    & 48 & \bf 0.0375 & \underline{ 0.1479} & $0.0381$ & $0.1487$ & $0.0381$ & $0.1485$ & \underline{0.0379} & \textbf{0.1478} & 0.0379 & 0.1479\\
    & 96 & \bf 0.0519 & \bf 0.1740 & $0.0551$ & $0.1810$ & $0.0554$ & $0.1808$ & \underline{0.0553} & \underline{0.1808} & 0.0555 & 0.1811\\
    & 192 & \bf 0.0676 & \bf 0.2013 & \underline{0.0680} & \underline{0.2019} & $0.0700$ & $0.2051$ & $0.0687$ & $0.2026$ & 0.0695 & 0.2041\\
    & 336 & \underline{0.0803} & \underline{0.2247} & $0.0805$ & $0.2252$ & \bf 0.0796 & \bf 0.2238 & $0.0806$ & $0.2254$ & 0.0803 & \underline{\bf 0.2244} \\
    & 720 & \bf 0.0776 & \bf 0.2224 & \underline{0.0808} & \underline{0.2271} & $0.0809$ & $0.2273$ & $0.0808$ & $0.2273$ & 0.0795 & 0.2255\\
    \hline
    \multirow{3}{*}{Illness} & 24 & $ 0.5881 $ & $ 0.5358 $ & \underline{0.5805} & \underline{0.5363} & $0.5845$ & $0.5376$ & \bf 0.5621 & \bf 0.5316 & \underline{\bf 0.5808} & 0.5464\\
    & 48 & $ 0.6858 $ & $ 0.6359 $ & \underline{0.6558} & \underline{0.6329} & $0.6813$ & $0.6535$ & \bf 0.6255 & \bf 0.5964 & \underline{\bf 0.6551} & \underline{\bf 0.6310}\\
    & 60 & \underline{0.6640} & \underline{0.6423} & $0.7481$ & $0.7029$ & \textbf{0.6506} & \bf 0.6265 & $0.7455$ & $0.6979$ & \underline{\bf 0.6513} & \underline{\bf 0.6270} \\
    \hline
    \multicolumn{2}{c}{}\\
    \end{tabular}
    \end{adjustbox}
    \caption{\textbf{State ablation study.} The table shows the forecasting performance across different episode progressions (non-sequential vs sequential), episode lengths (dynamic vs static), and state encodings (positional encoding vs zero padding). The best performance is highlighted in \textbf{bold}, while the second-best performance is \underline{underlined}. Additionally, for the comparison between positional encoding and zero padding, the outperformed performance of zero padding is \underline{\textbf{highlighted}}.} 
    \label{tab: state}
\end{table*}
We conduct an ablation study on episode progression and state encoding. First, the progression of the episode is divided into two approaches: a non-sequential approach based on random sampling and a sequential approach following the conventional time axis order. Next, the composition of episode length is categorized into a dynamic approach with random sampling and a static approach using a fixed window. Note that the static length is set to 1500 for Exchange and ETTh1, and 200 for Illness. As shown in Table `State', it is evident that the non-sequential episodes with dynamic length exhibit robust performance. Following this, the traditional approach, sequential episodes with static length, shows the second most robust performance. This confirms that DTF-net achieves the best performance while preventing overfitting issues, in contrast to the sequential episode with static length method that exhibits overfitted results.

Next, positional encoding achieves superior performance compared to zero padding, except for the Illness dataset. For the Illness dataset, zero padding performs better due to its relatively short look-back horizon of 104, making simple zero padding more effective compared to Exchange and ETTh. However, as the look-back horizon for forecasting increases, reaching 336 in Exchange and ETTh1, positional encoding proves to be more effective.

\newpage
\subsection{Ablation Study on Reward with Seed Test}
\label{app: reward}

\begin{table*}[ht]
    \centering
    \begin{adjustbox}{width=1\textwidth}
    \begin{tabular}{c|c||cc||cc|cc|cc|cc|cc|cc|cc|cc|cc|cc}  
    \hline \multicolumn{2}{c||}{Seed} & \multicolumn{2}{c||}{ 2023 } & \multicolumn{2}{c|}{ 52 } & \multicolumn{2}{c|}{ 454 } & \multicolumn{2}{c|}{ 470 } & \multicolumn{2}{c|}{ 515 } & \multicolumn{2}{c|}{ 695 } & \multicolumn{2}{c|}{ 1561 } & \multicolumn{2}{c|}{ 1765 } & \multicolumn{2}{c|}{ 1953 } & \multicolumn{2}{c|}{ 2021 } & \multicolumn{2}{c}{ 2022 } \\
    \hline \multicolumn{2}{c||}{Metric} & MSE & MAE & MSE & MAE & MSE & MAE & MSE & MAE & MSE & MAE & MSE & MAE & MSE & MAE & MSE & MAE & MSE & MAE & MSE & MAE & MSE & MAE \\
    \hline 

    \multirow{6}{*}{Exchange} & 24 & $ 0.0250 $ & $ 0.1198 $ & $ 0.0265 $ & $ 0.1236 $ & $ 0.0256 $ & $ 0.1211 $ & $ 0.0256 $ & $ 0.1220 $ & $ 0.0258 $ & $ 0.1209 $ & $ 0.0267 $ & $ 0.1237 $ & $ 0.0261 $ & $ 0.1226 $ & $ 0.0253 $ & $ 0.1203 $ & $ 0.0267 $ & $ 0.1239 $ & $ 0.0263 $ & $ 0.1227 $ & $ 0.0265 $ & $ 0.1231 $  \\
    & 48 & $ 0.0487 $ & $ 0.1658 $ & $ 0.0499 $ & $ 0.1682 $ & $ 0.0498 $ & $ 0.1677 $ & $ 0.0503 $ & $ 0.1691 $ & $ 0.0492 $ & $ 0.1658 $ & $ 0.0492 $ & $ 0.1674 $ & $ 0.0500 $ & $ 0.1681 $ & \bf 0.0480 & \bf 0.1655 & $ 0.0493 $ & $ 0.1680 $ & \bf 0.0480 & \bf 0.1650 & $ 0.0510 $ & $ 0.1703 $  \\
    & 96 & $ 0.0983 $ & $ 0.2349 $ & $ 0.1007 $ & $ 0.2363 $ & $ 0.0989 $ & $ 0.2349 $ & \bf 0.0970 & \bf 0.2322 & $ 0.0985 $ & \bf 0.2322 & $ 0.0982 $ & $ 0.2356 $ & $ 0.0992 $ & $ 0.23540 $ & $ 0.0994 $ & $ 0.2360 $ & $ 0.0982 $ & $ 0.2350 $ & $ 0.1007 $ & $ 0.2335 $ & $ 0.1004 $ & $ 0.2368 $  \\
    & 192 & $ 0.1983 $ & $ 0.3583 $ & $ 0.2096 $ & $ 0.3673 $ & $ 0.2004 $ & $ 0.3562 $ & $ 0.2023 $ & $ 0.3565 $ & $ 0.2115 $ & $ 0.3751 $ & $ 0.2009 $ & $ 0.3567 $ & \bf 0.1978 & $ 0.3595 $ & $ 0.2063 $ & $ 0.3667 $ & \bf 0.1904 & \bf 0.3513 & $ 0.2056 $ & $ 0.3649 $ & \bf 0.1929 & \bf 0.3512  \\
    & 336 & $ 0.3160 $ & $ 0.4561 $ & $ 0.4085 $ & $ 0.4939 $ & $ 0.3311 $ & $ 0.4712 $ & $ 0.3288 $ & $ 0.4678 $ & $ 0.3409 $ & $ 0.4716 $ & $ 0.3180 $ & $ 0.4467 $ & $ 0.3199 $ & $ 0.4587 $ & $ 0.3559 $ & $ 0.4874 $ & $ 0.3333 $ & $ 0.4626 $ & \bf 0.3124 & \bf 0.4578 & $ 0.3606 $ & $ 0.4768 $  \\
    & 720 & $ 0.7933 $ & $ 0.6874 $ & \bf 0.7583 & $ 0.6847 $ & $ 0.8022 $ & $ 0.7083 $ & $ 0.8888 $ & $ 0.7331 $ & $ 1.0462 $ & $ 0.7938 $ & $ 0.9455 $ & $ 0.7649 $ & $ 0.9338 $ & $ 0.7445 $ & $ 1.0001 $ & $ 0.7788 $ & $ 0.8769 $ & $ 0.8769 $ & $ 0.9480 $ & $ 0.7534 $ & \bf 0.7889 & $ 0.6985 $  \\
    \hline
    \multirow{6}{*}{ETTh1} & 24 & $ 0.0253 $ & $ 0.1205 $ & $ 0.0258 $ & $ 0.1219 $ & $ 0.0259 $ & $ 0.1218 $ & $ 0.0266 $ & $ 0.1234 $ & $ 0.0259 $ & $ 0.1222 $ & \bf 0.0250 & \bf 0.1197 & $ 0.0261 $ & $ 0.1224 $ & $ 0.0264 $ & $ 0.1235 $ & $ 0.0263 $ & $ 0.1233 $ & $ 0.0258 $ & $ 0.1214 $ & $ 0.0260 $ & $ 0.1221 $  \\
    & 48 & $ 0.0375 $ & $ 0.1479 $ & $ 0.0381 $ & $ 0.1485 $ & $ 0.0383 $ & $ 0.1489 $ & $ 0.0388 $ & $ 0.1501 $ & $ 0.0404 $ & $ 0.1546 $ & $ 0.0388 $ & $ 0.1497 $ & $ 0.0391 $ & $ 0.1507 $ & $ 0.0389 $ & $ 0.1505 $ & \bf 0.0370 & \bf 0.1467 & $ 0.0391 $ & $ 0.1508 $ & $ 0.0386 $ & $ 0.1490 $  \\
    & 96 & $ 0.0519 $ & $ 0.1740 $ & $ 0.0528 $ & $ 0.1763 $ & $ 0.0561 $ & $ 0.1824 $ & $ 0.0532 $ & $ 0.1774 $ & $ 0.0536 $ & $ 0.1774 $ & $ 0.0540 $ & $ 0.1780 $ & $ 0.0544 $ & $ 0.1789 $ & $ 0.0525 $ & $ 0.1755 $ & $ 0.0531 $ & $ 0.1768 $ & $ 0.0539 $ & $ 0.1777 $ & $ 0.0536 $ & $ 0.1783 $  \\
    & 192 & $ 0.0676 $ & $ 0.2013 $ & $ 0.0697 $ & $ 0.2041 $ & $ 0.0683 $ & $ 0.2036 $ & $ 0.0704 $ & $ 0.2057 $ & $ 0.0698 $ & $ 0.2048 $ & $ 0.0703 $ & $ 0.2054 $ & $ 0.0681 $ & $ 0.2024 $ & $ 0.0695 $ & $ 0.2034 $ & $ 0.0695 $ & $ 0.2049 $ & $ 0.0694 $ & $ 0.2044 $ & $ 0.0686 $ & $ 0.2032 $  \\
    & 336 & $ 0.0803 $ & $ 0.2247 $ & $ 0.0856 $ & $ 0.2325 $ & \bf 0.0781 & \bf 0.2231 & $ 0.0818 $ & $ 0.2272 $ & $ 0.0834 $ & $ 0.2285 $ & $ 0.0833 $ & $ 0.2291 $ & \bf 0.0801 & \bf 0.2247 & \bf 0.0792 & \bf 0.2232 & $ 0.0815 $ & $ 0.2266 $ & $ 0.0819 $ & $ 0.2278 $ & $ 0.0807 $ & $ 0.2253 $  \\
    & 720 & $ 0.0776 $ & $ 0.2224 $ & $ 0.0784 $ & $ 0.2239 $ & $ 0.0823 $ & $ 0.2288 $ & $ 0.0820 $ & $ 0.2285 $ & $ 0.0832 $ & $ 0.2304 $ & $ 0.0820 $ & $ 0.2278 $ & $ 0.0813 $ & $ 0.2278 $ & $ 0.0814 $ & $ 0.2282 $ & $ 0.0792 $ & $ 0.2255 $ & $ 0.0802 $ & $ 0.2252 $ & $ 0.0805 $ & $ 0.2262 $  \\
    \hline
    \multirow{3}{*}{Illness} & 24 & $ 0.5881 $ & $ 0.5358 $ & $ 0.6275 $ & $ 0.5617 $ & $ 0.6100 $ & $ 0.5498 $ & $ 0.6684 $ & $ 0.5718 $ & $ 0.6285 $ & $ 0.5689 $ & \bf 0.5647 & \bf 0.5117 & \bf 0.5764 & \bf 0.5310 & $ 0.6112 $ & $ 0.5424 $ & \bf 0.5476 & \bf 0.5303 & $ 0.7151 $ & $ 0.6080 $ & $ 0.6570 $ & $ 0.5621 $  \\
    & 48 & $ 0.6858 $ & $ 0.6359 $ & $ 0.7178 $ & $ 0.6531 $ & \bf 0.6534 & \bf 0.6170 & $ 0.7407 $ & $ 0.6822 $ & $ 0.7241 $ & $ 0.6626 $ & \bf 0.5881 & \bf 0.5655 & \bf 0.6829 & \bf 0.6604 & \bf 0.6343 & \bf 0.6021 & \bf 0.5991 & \bf 0.5642 & $ 0.7104 $ & $ 0.6532 $ & $ 0.7159 $ & $ 0.6435 $  \\
    & 60 & $ 0.6640 $ & $ 0.6423 $ & $ 0.6682 $ & $ 0.6448 $ & $ 0.7822 $ & $ 0.7338 $ & $ 0.7314 $ & $ 0.6879 $ & \bf 0.6492 & \bf 0.6290 & $ 0.7194 $ & $ 0.6805 $ & $ 0.7020 $ & $ 0.6607 $ & \bf 0.6526 & \bf 0.6399 & $ 0.7493 $ & $ 0.6955 $ & $ 0.6769 $ & $ 0.6408 $ & \bf 0.6479 & \bf 0.6196  \\
    \hline
    \multicolumn{2}{c}{}\\
    \end{tabular}
    \label{tab: seed_result_interval_reward}
    \end{adjustbox}
    \bigskip
    \begin{adjustbox}{width=0.5\textwidth}
    \begin{tabular}{c|c|c|c|c|c|c|c}
        \hline
         var / length & 24 & 48 & 60 & 96 & 192 & 336 & 720\\
         \hline
         Exchange & $3e-7$ & $9e-7$ & - & $1e-6$ & $5e-5$ & $8e-4$ & $9e-3$ \\
         \hline
         ETTh1 & $1e-7$ & $7e-7$ & - & $1e-6$ & $6e-7$ & $5e-6$ & $2e-6$  \\
         \hline
         Illness & $3e-3$ & $3e-3$ & $2e-3$ & - & - & - & - \\
         \hline
    \end{tabular}
    \end{adjustbox}
    \caption{\textbf{Equal interval reward sampling.} The table presents the forecasting performance across different seeds with rewards sampled at equal intervals. Results from the seed that demonstrated superior performance, using the 2023 seed with a random interval as the reference result, are highlighted in \textbf{bold}. The second table illustrates the performance variance of the seed tests based on MSE.} 
    \label{tab:interval}
\end{table*}

\begin{table*}[ht]
    \centering
    \begin{adjustbox}{width=1\textwidth}
    \begin{tabular}{c|c||cc||cc|cc|cc|cc|cc|cc|cc|cc|cc|cc}  
    \hline \multicolumn{2}{c||}{Seed} & \multicolumn{2}{c||}{ 2023 } & \multicolumn{2}{c|}{ 52 } & \multicolumn{2}{c|}{ 454 } & \multicolumn{2}{c|}{ 470 } & \multicolumn{2}{c|}{ 515 } & \multicolumn{2}{c|}{ 695 } & \multicolumn{2}{c|}{ 1561 } & \multicolumn{2}{c|}{ 1765 } & \multicolumn{2}{c|}{ 1953 } & \multicolumn{2}{c|}{ 2021 } & \multicolumn{2}{c}{ 2022 } \\
    \hline \multicolumn{2}{c||}{Metric} & MSE & MAE & MSE & MAE & MSE & MAE & MSE & MAE & MSE & MAE & MSE & MAE & MSE & MAE & MSE & MAE & MSE & MAE & MSE & MAE & MSE & MAE \\
    \hline 

    \multirow{6}{*}{Exchange} & 24  & 0.0250 & 0.1198 & 0.0265 & 0.1237 & 0.0251 & 0.1198 & 0.0259 & 0.1228 & 0.0250 & 0.1186 & 0.0254 & 0.1207 & 0.0254 & 0.1205 & 0.0257 & 0.1210 & 0.0269 & 0.1244 & 0.0263 & 0.1230 & 0.0261 & 0.1218 \\
    & 48  & 0.0487 & 0.1658 & 0.0491 & 0.1666 & 0.0498 & 0.1678 & 0.0502 & 0.1682 & 0.0492 & 0.1658 & 0.0491 & 0.1665 & 0.0497 & 0.1670 & 0.0479 & 0.1652 & 0.0495 & 0.1682 & 0.0485 & 0.1665 & 0.0498 & 0.1683 \\
    & 96  & 0.0983 & 0.2349 & 0.1003 & 0.2356 & 0.1009 & 0.2348 & 0.0970 & 0.2322 & 0.0968 & 0.2308 & 0.0952 & 0.2323 & 0.0985 & 0.2345 & 0.0991 & 0.2355 & 0.0996 & 0.2366 & 0.1006 & 0.2343 & 0.1004 & 0.2366 \\
    & 192 & 0.1983 & 0.3583 & 0.2091 & 0.3652 & 0.2019 & 0.3575 & 0.1994 & 0.3538 & 0.2130 & 0.3760 & 0.2028 & 0.3579 & 0.1990 & 0.3593 & 0.2093 & 0.3693 & 0.1907 & 0.3515 & 0.2055 & 0.3648 & 0.1924 & 0.3508 \\
    & 336 & 0.3160 & 0.4561 & 0.4040 & 0.4926 & 0.3329 & 0.4728 & 0.3284 & 0.4675 & 0.3409 & 0.4719 & 0.3202 & 0.4480 & 0.3193 & 0.4581 & 0.3571 & 0.4880 & 0.3334 & 0.4629 & 0.3122 & 0.4585 & 0.3638 & 0.3638 \\
    & 720 & 0.7933 & 0.6874 & 0.7600 & 0.6855 & 0.7693 & 0.6920 & 0.8888 & 0.7332 & 1.0404 & 0.7913 & 0.9442 & 0.7636 & 0.9398 & 0.7483 & 0.9685 & 0.7745 & 0.8746 & 0.7249 & 0.9457 & 0.7529 & 0.7856 & 0.6955  \\
    \hline
    \multirow{6}{*}{ETTh1} & 24  & 0.0253 & 0.1205 & 0.0257 & 0.1214 & 0.0258 & 0.1218 & 0.0268 & 0.1235 & 0.0257 & 0.1218 & \bf 0.0250 & \bf 0.1196 & 0.0261 & 0.1225 & 0.0260 & 0.1222 & 0.0254 & 0.1203 & 0.0258 & 0.1215 & 0.0261 & 0.1222 \\
    & 48  & 0.0375 & 0.1479 & 0.0381 & 0.1485 & 0.0386 & 0.1493 & 0.0396 & 0.1513 & 0.0378 & 0.1488 & 0.0389 & 0.1499 & 0.0391 & 0.1506 & 0.0387 & 0.1502 & 0.0370 & 0.1470 & 0.0388 & 0.1502 & 0.0385 & 0.1490 \\
    & 96  & 0.0519 & 0.1740 & 0.0529 & 0.1765 & 0.0557 & 0.1819 & 0.0535 & 0.1779 & 0.0537 & 0.1778 & 0.0525 & 0.1751 & 0.0543 & 0.1789 & 0.0526 & 0.1757 & 0.0525 & 0.1760 & 0.0538 & 0.1774 & 0.0536 & 0.1783 \\
    & 192 & 0.0676 & 0.2013 & 0.0698 & 0.2040 & 0.0709 & 0.2067 & 0.0717 & 0.2081 & 0.0698 & 0.2049 & 0.0697 & 0.2046 & 0.0680 & 0.2023 & 0.0696 & 0.2034 & 0.0703 & 0.2063 & 0.0695 & 0.2042 & 0.0690 & 0.2040 \\
    & 336 & 0.0803 & 0.2247 & 0.0845 & 0.2314 & \bf 0.0783 & \bf 0.2229 & 0.0818 & 0.2272 & 0.0825 & 0.2277 & 0.0844 & 0.2302 & 0.0800 & 0.2251 & 0.0822 & 0.2268 & 0.0806 & 0.2256 & 0.0820 & 0.2271 & 0.0807 & 0.2250 \\
    & 720 & 0.0776 & 0.2224 & 0.0782 & 0.2234 & 0.0808 & 0.2267 & 0.0826 & 0.2292 & 0.0823 & 0.2292 & 0.0790 & 0.2237 & 0.0808 & 0.2272 & 0.0801 & 0.2253 & 0.0805 & 0.2271 & 0.0804 & 0.2254 & 0.0800 & 0.2253  \\
    \hline
    \multirow{3}{*}{Illness} & 24 & 0.5881 & 0.5358 & 0.6157 & 0.5495 & 0.6326 & 0.5689 & 0.6237 & 0.5672 & 0.7393 & 0.6669 & \bf 0.5641 & \bf 0.5293 & 0.6009 & 0.5684 & 0.6278 & 0.5575 & 0.6121 & 0.5471 & 0.6330 & 0.5590 & 0.6681 & 0.5958 \\
    & 48 & 0.6858 & 0.6359 & 0.7181 & 0.6537 & 0.7076 & 0.6407 & 0.7800 & 0.7061 & 0.7205 & 0.6726 & \bf 0.6398 & \bf 0.5907 & 0.6958 & 0.6621 & \bf 0.6363 & \bf 0.6025 & \bf 0.6254 & \bf 0.5743 & \bf 0.6340 & \bf 0.6171 & 0.7155 & 0.6495 \\
    & 60 & 0.6640 & 0.6423 & \bf 0.6542 & \bf 0.6329 & 0.7767 & 0.7326 & 0.7394 & 0.6881 & 0.6728 & 0.6501 & 0.7359 & 0.6929 & 0.7140 & 0.6658 & 0.6626 & 0.6460 & 0.7291 & 0.6829 & \bf 0.6598 & \bf 0.6236 & \bf 0.6483 & \bf 0.6203  \\
    \hline
    \multicolumn{2}{c}{}\\
    \end{tabular}
    \label{tab: seed_result_random_reward}
    \end{adjustbox}
    \bigskip
    \begin{adjustbox}{width=0.5\textwidth}
    \begin{tabular}{c|c|c|c|c|c|c|c}
        \hline
         var / length & 24 & 48 & 60 & 96 & 192 & 336 & 720\\
         \hline
         Exchange & $4e-7$ & $5e-7$ & - & $4e-7$ & $5e-7$ & $8e-4$ & $9e-3$\\
         \hline
         ETTh1 & $2e-7$ & $5e-7$ & - & $1e-6$ & $1e-6$ & $4e-6$ & $2e-6$\\
         \hline
         Illness & $2e-3$ & $3e-3$ & $2e-3$ &  - & - & - & -  \\
         \hline
    \end{tabular}
    \end{adjustbox}
    \caption{\textbf{Random interval reward sampling.} The table presents the forecasting performance across different seeds with rewards sampled at random intervals. Results from the seed that demonstrated superior performance, using the 2023 seed with a random interval as the reference result, are highlighted in \textbf{bold}. The second table illustrates the performance variance of the seed tests based on MSE.} 
    \label{tab:random}
\end{table*}

We conduct experiments under two conditions to assess the performance of DTF-net using different reward sampling methods with 10 randomly selected seeds: 1) sampling rewards at equal intervals (Table \ref{tab:interval}), and 2) sampling rewards at random intervals (Table \ref{tab:random}). As shown in the tables, there are no significant disparities between the two interval reward sampling methods. However, concerning the seed test, as the prediction horizon increases, the variance of performance of the seed test also increases. This limitation is inherent to RL's sensitivity to hyperparameters, including seeds. Nevertheless, it is important to note that this sensitivity can also be advantageous, as optimal performance can be achieved by identifying the appropriate combination of hyperparameters.

\clearpage
\section{Limitation of DTF-net}

\subsection{DTF-net on Stationary TSF: Weather and Traffic Dataset}
\begin{table*}[ht]
    \centering
    \begin{adjustbox}{width=1\textwidth}
    \begin{tabular}{c|c|cc|cc|cc|cc|cc|cc|cc}  
    \hline \multicolumn{2}{c|}{Methods} & \multicolumn{2}{c|}{ DTF-Linear (ours) } & \multicolumn{2}{c|}{ $\ell_1 (\lambda =0.1)$-Linear } & \multicolumn{2}{c|}{ NLinear } & \multicolumn{2}{c|}{ DLinear } & \multicolumn{2}{c|}{ FEDformer-f } & \multicolumn{2}{c|}{ FEDformer-w } & \multicolumn{2}{c}{ Autoformer } \\
    \hline \multicolumn{2}{c|}{Metric} & MSE & MAE & MSE & MAE & MSE & MAE & MSE & MAE & MSE & MAE & MSE & MAE & MSE & MAE \\
    \hline 
    % \cline{3-8}
    \multirow{6}{*}{Weather} & 24 & 0.0004 & 0.0143 & 0.0004 & 0.0133 & \underline{$0.0004$} & \underline{$0.0128$} & $0.0019$ & $0.0323$ & $0.0272$ & $0.1263$ & $0.0217$ & $0.0217$ & $0.0133$ & $0.0957$ \\
    & 48 & 0.0008 & 0.0208 & \underline{\textit{$0.0008$}} & \underline{\textit{$0.0188$}} & \underline{$0.0007$} & \underline{$0.0190$} & $0.0043$ & $0.0513$ & $0.0053$ & $0.0586$ & $0.0055$ & $0.0599$ & $0.0115$ & $0.0850$  \\
    & 96 & 0.0010 & 0.0236 & \underline{\textit{0.0010}} & \underline{\textit{0.0227}} & \underline{$0.0010$} & \underline{$0.0233$} & $0.0047$ & $0.0543$ & $0.0096$ & $0.0770$ & $0.0055$ & $0.0593$ & $0.0094$ & $0.0769$  \\
    & 192 & \bf 0.0012 & \bf 0.0253 & $0.0012$ & $0.0257$ & \underline{$0.0012$} & \underline{$0.0261$} & {$0.0054$} & $0.0591$ & $0.0048$ & $0.0558$ & $0.0048$ & $0.0559$ & $0.0055$ & $0.0570$ \\
    & 336 & \bf 0.0014 & \bf 0.0277 & $0.0014$ & $0.0279$ & \underline{0.0014} & \underline{0.0278} & {$0.0064$} & {$0.0664$} & $0.0049$ & $0.0554$ & $0.0049$ & $0.0552$ & $0.0082$ & $0.0683$ \\
    & 720 & \bf 0.0019 & \bf 0.0318 & $0.0019$ & $0.0323$ & \underline{$0.0019$} & \underline{$0.0329$} & $0.0066$ & $0.0679$ & $0.0036$ & $0.0479$ & $0.0036$ & $0.0478$ & $0.0055$ &$0.0561$ \\
    \hline
    \multirow{6}{*}{Traffic} & 24 & \bf 0.1155 & 0.1963 & $0.1248$ & $0.2174$ & $0.1159$ & \underline{$0.1962$} & $0.1166$ & $0.1986$ & $0.1526$ & $0.2535$ & $0.1506$ & $0.2432$ & $0.2279$ & $0.3461$ \\
    & 48 & $0.1251$ & $0.2110$ & $0.1327$ & $0.2240$ & \underline{$0.1214$} & \underline{$0.2002$} & $0.1228$ & $0.3505$ & $0.1729$ & $0.2772$ & $0.1803$ & $0.2759$ & $0.2523$ & $0.3666$  \\
    & 96 & 0.1391 & 0.2283 & 0.1402 & 0.2322 & \underline{$0.1282$} & \underline{$0.2074$} & $0.1300$ & $0.2114$ & $0.1890$ & $0.2884$ & $0.1933$ & $0.2872$ & $0.2550$ & $0.3665$  \\
    & 192 & 0.1389 & 0.2263 & $0.1429$ & $0.2354$ & \underline{$0.1328$} & \underline{$0.2132$} & $0.1331$ & $0.2151$ & $0.1901$ & $0.2936$ & $0.1955$ & $0.2978$ & $0.2531$ & $0.3594$ \\
    & 336 & 0.1619 & 0.2629 & {0.1419} & {0.2377} & \underline{$0.1301$} & \underline{$0.2163$} & {$0.1331$} & {$0.2213$} & $0.1980$ & $0.3073$ & $0.2000$ & $0.3092$ & $0.2965$ & $0.3926$ \\
    & 720 & 0.1548 & 0.2518 & 0.1550 & 0.2516 & \underline{$0.1423$} & \underline{$0.2283$} & $0.1455$ & $0.2349$ & $0.2601$ & $0.3469$ & $0.2634$ & $0.3474$ & $0.3935$ & $0.4562$  \\
    \hline
    \end{tabular}
    \end{adjustbox}
    \caption{\textbf{DTF-net on stationary TSF.} We conduct TSF experiments using two stationary datasets: Weather and Traffic. Performance is evaluated using MSE and MAE, where lower values indicate better performance. In the following results, the best-performing models using DTF-net are highlighted in \textbf{bold}, and models using $\ell_1$ trend filtering are highlighted in \underline{\textit{italic}}. To compare the results, the best-performing models using the original data are \underline{underlined}.} 
    \label{tab:stationary}
\end{table*}

DTF-net is specifically designed to tackle the trend filtering problem, focusing on capturing abrupt changes driven by extreme values. Therefore, DTF-net may not be the most suitable choice for handling stationary datasets. To support this assertion, we observe that both DTF-net and $\ell_1$ trend filtering exhibit suboptimal performance in terms of TSF when applied to stationary datasets. This suggests that the emphasis on trend filtering, particularly in capturing abrupt changes, may impact the performance of TSF negatively. Nevertheless, as indicated by the results in Table \ref{tab:TSF_main} and Figure \ref{fig:extreme value study}, we substantiate that DTF-net significantly improves TSF performance when dealing with non-stationary and complex datasets.

\subsection{Multivariate Trend Filtering and Time Series Forecasting}
DTF-net is employed for univariate trend filtering on features derived from multivariate time series data. In this context, RL leverages multidimensional aspects to perform trend filtering specifically on the target variable. It's worth noting that this study does not explicitly address multivariate trend filtering for all dimensions within the time series data. However, we believe that achieving this is feasible through the multi-discrete action prediction capability of the PPO algorithm \cite{schulman2017proximal}.

\subsection{Stability of DTF-net}
\begin{figure*}[h]
\centering
    \subfloat[ETTh1 Reward]{\includegraphics[clip, trim = 0 0 3.8cm 0, width=0.4\textwidth]{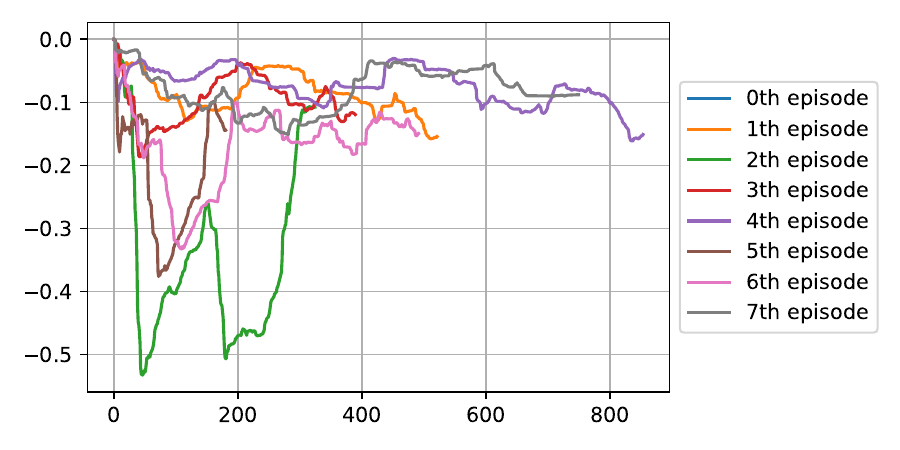}}
    \subfloat[EXC Reward]{\includegraphics[width=0.53\textwidth]{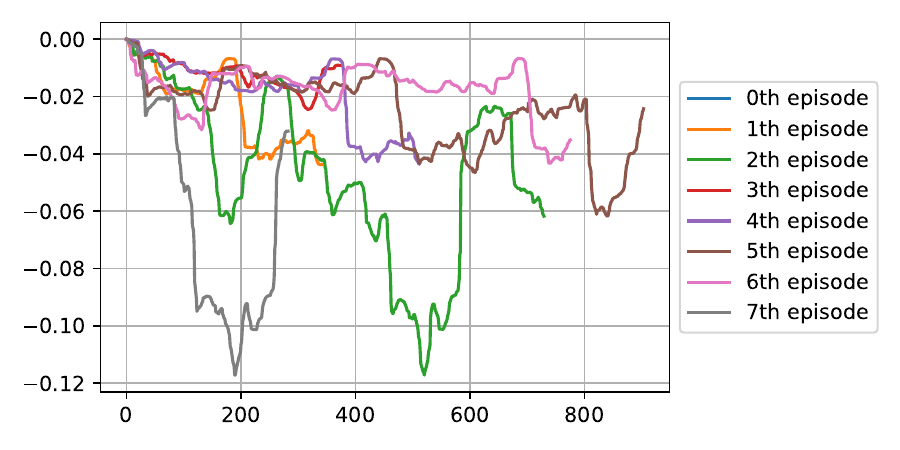}}
    \caption{The figure shows rewards obtained from dynamically segmented sub-sequences across different episodes from ETTh1 and Exchange Rate (EXC). Since DTF-net uses the forecasting cost function as a reward, the reward tends to be unstable.}
\end{figure*}
Regarding stability and convergence, in each episode, there is a tendency for cumulative reward to increase, but due to the nature of forecasting MSE, the reward can be unstable depending on the characteristics of the input data batch. While acknowledging the degradation in complexity and stability, DTF-net demonstrates clear advantages in extracting dynamic trends.

\end{document}